\UseRawInputEncoding
\documentclass{article}

\usepackage[preprint]{neurips_2024}
\usepackage{hyperref}
\usepackage{url}
\usepackage{comment}
\usepackage{enumitem}
\usepackage{booktabs}       
\usepackage{amsfonts}       
\usepackage{nicefrac}       
\usepackage{microtype}      
\usepackage{wrapfig}
\usepackage{lipsum}
\usepackage{amssymb,amsfonts}
\usepackage{amsthm}
\usepackage{graphicx}
\usepackage{amsmath}
\usepackage{algorithm,algorithmic}
\usepackage{multirow}
\usepackage{color}

\usepackage[utf8]{inputenc} 
\usepackage[T1]{fontenc}    
\usepackage{hyperref}       
\usepackage{url}            
\usepackage{booktabs}       
\usepackage{amsfonts}       
\usepackage{nicefrac}       
\usepackage{microtype}      
\usepackage{xcolor}         

\usepackage{diagbox}
\usepackage{multirow}
\usepackage{multicol}
\usepackage{graphicx}
\usepackage{subcaption} 
\usepackage{algorithm}
\usepackage{algorithmic}
\usepackage{longtable}
\usepackage{supertabular}

\title{Graph Transductive Defense: a Two-Stage Defense for Graph Membership Inference Attacks}

\author{%
  Peizhi Niu \thanks{Equal contribution. Work done during the internship of Peizhi Niu at University of Illinois Urbana-Champaign.}   \\
  Shanghai Jiao Tong University \\
  \texttt{npz$\_$liekon@sjtu.edu.cn} \\
  \And
  Chao Pan \footnotemark[1] \\
  University of Illinois Urbana-Champaign \\
  \texttt{chaopan2@illinois.edu} \\
  \AND
  Siheng Chen \\
  Shanghai Jiao Tong University $\&$ Shanghai AI Laboratory \\
  \texttt{sihengc@sjtu.edu.cn} \\
  \And
  Olgica Milenkovic\\
  University of Illinois Urbana-Champaign \\
  \texttt{milenkov@illinois.edu} \\
}

\begin{document}

\maketitle

\begin{abstract}
\label{abstract}

  Graph neural networks (GNNs) have become instrumental in diverse real-world applications, offering powerful graph learning capabilities for tasks such as social networks and medical data analysis. Despite their successes, GNNs are vulnerable to adversarial attacks, including membership inference attacks (MIA), which threaten privacy by identifying whether a record was part of the model's training data. While existing research has explored MIA in GNNs under graph inductive learning settings, the more common and challenging graph transductive learning setting remains understudied in this context. This paper addresses this gap and proposes an effective two-stage defense, Graph Transductive Defense (GTD), tailored to graph transductive learning characteristics. The gist of our approach is a combination of a train-test alternate training schedule and flattening strategy, which successfully reduces the difference between the training and testing loss distributions. Extensive empirical results demonstrate the superior performance of our method (a decrease in attack AUROC by $9.42\%$ and an increase in utility performance by $18.08\%$ on average compared to LBP), highlighting its potential for seamless integration into various classification models with minimal overhead.
\end{abstract}


\maketitle
\vspace{-0.1in}
\section{Introduction}
\vspace{-0.05in}
\label{sec:intro}
Graph neural networks (GNNs) have emerged as a series of competent graph learning methods for diverse real-world scenarios, ranging from social networks and recommendation systems to biological data analysis~\citep{wu2021a,zhang2024trustworthy}. For example, GNNs have been shown to be powerful in improving personalized search and recommendations for customers on e-commerce platforms (e.g., AliGraph at Alibaba~\citep{zhu2019aligraph} and GIANT at Amazon~\citep{chien2021node}) and social networks (e.g., PinnerSage at Pinterest~\citep{pal2020pinnerSage} and LiGNN at LinkedIn~\citep{borisyuk2024lignn}). Compared to traditional deep learning methods, which assume data originating from Euclidean space, GNNs can make full use of the additional graph topology information between data points through specialized operations (i.e., graph convolutions). These operations allow GNNs to generate more informative embeddings that are better suited for downstream tasks.

Despite the success of GNNs, they have also been shown to be prone to various adversarial attacks~\citep{sun2023adversarial}, including membership inference attacks (MIA)~\citep{shokri2017membership,hu2022membership}. MIA involves determining whether a given record (i.e., some data points) is part of the training dataset used to build a specific target model, given the model itself, the record and information about the dataset. Typically, MIA uses prediction logits of shadow model to train attack models, where the shadow training data is obtained either by inferencing the target model, or the attacker directly have access to a potentially noisy version of the original training dataset. After the attack, if the attacker knows that a record was used to train a particular model, it implies an information leakage through the model. For example, if a GNN is trained on nodes belonging to a private group (e.g., support group for sensitive medical issues) within a large social network, successful MIA to this GNN could reveal patient identity information and lead to severe privacy breaches. 

\begin{figure*}[tbp]
    \centering
    \begin{subfigure}[b]{0.49\textwidth}
        \centering
        \includegraphics[width=\linewidth]{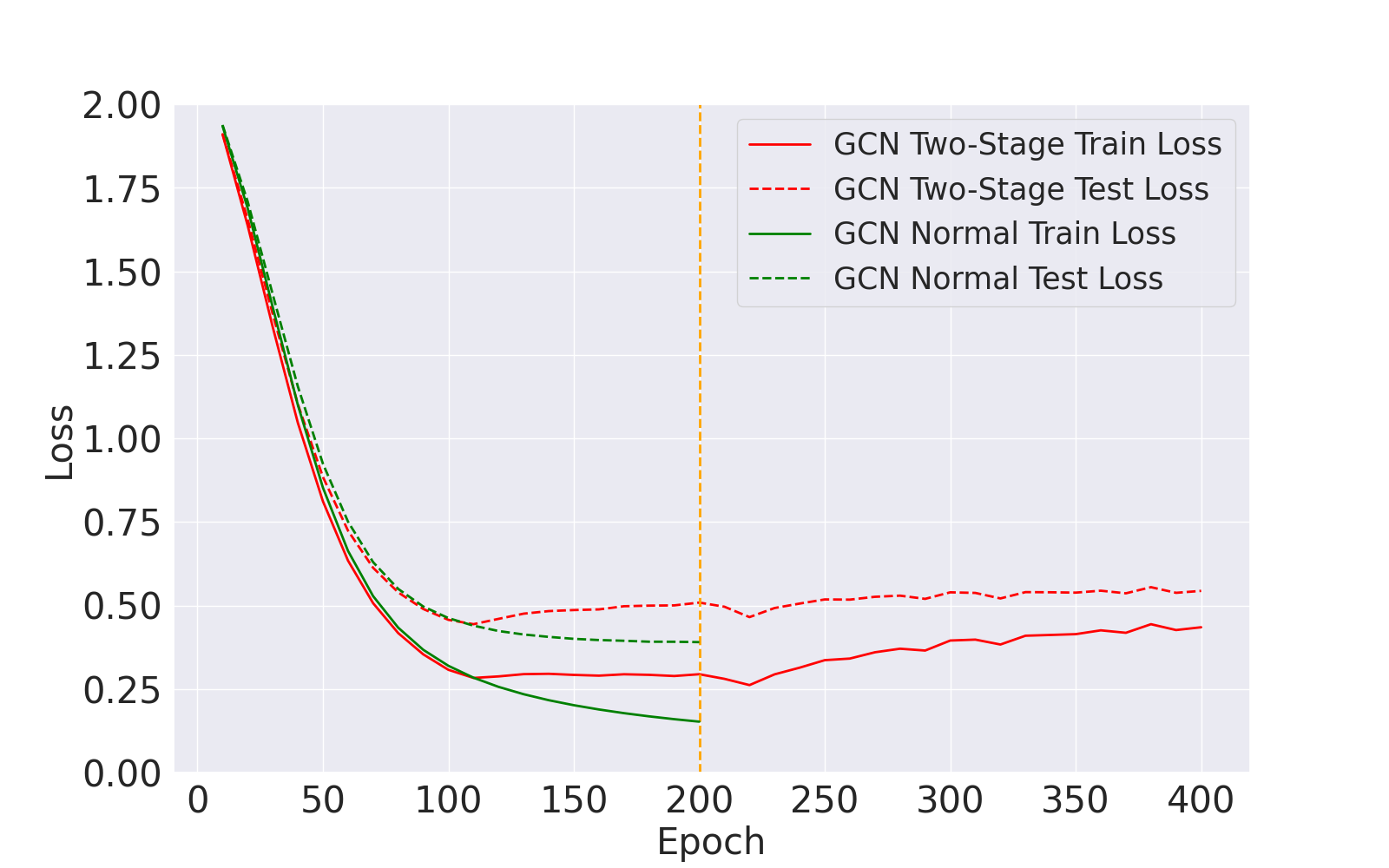}
        \caption{}
    \end{subfigure}
    \begin{subfigure}[b]{0.49\textwidth}
        \includegraphics[width=\linewidth]{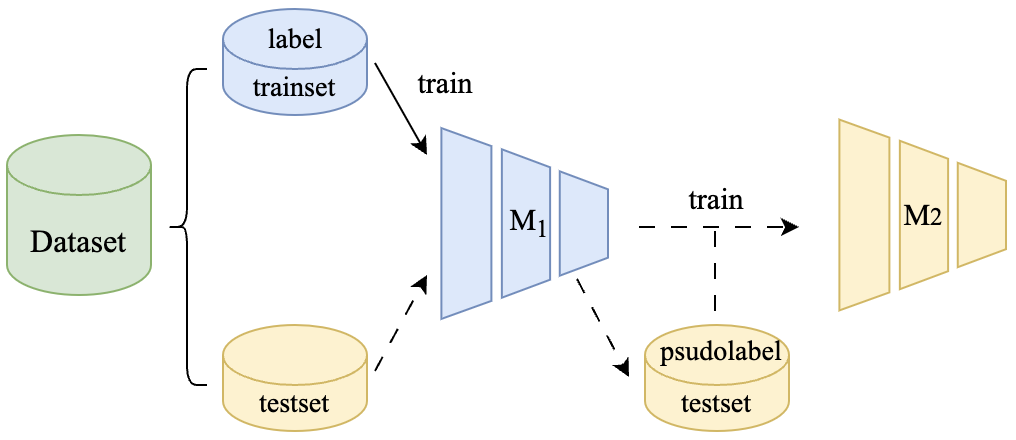}
        \caption{}
    \end{subfigure}
    \begin{subfigure}[b]{0.49\textwidth}
        \centering
        \includegraphics[width=\linewidth]{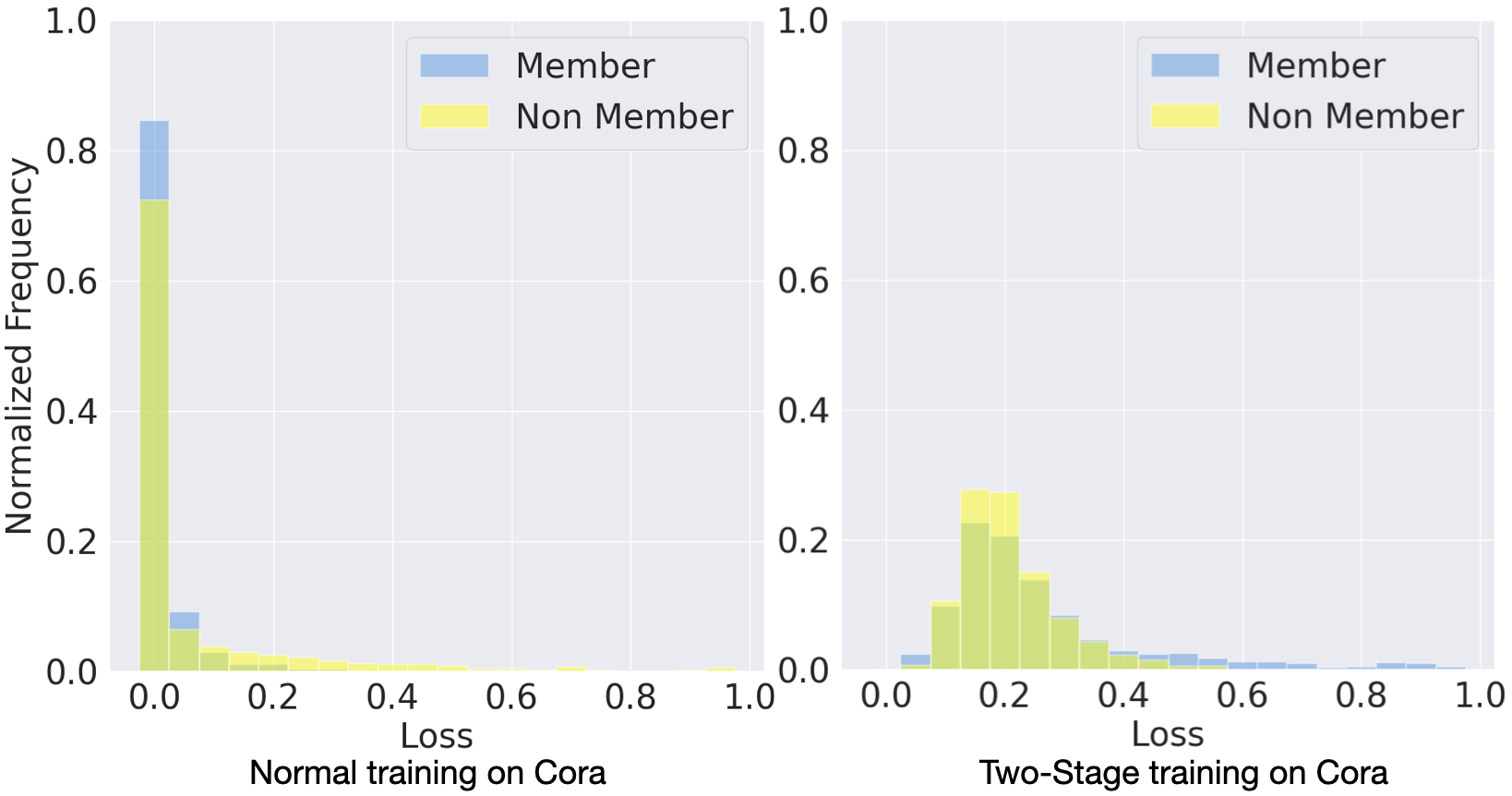}
        \caption{}
    \end{subfigure}
    \begin{subfigure}[b]{0.49\textwidth}
        \centering
        \includegraphics[width=\linewidth]{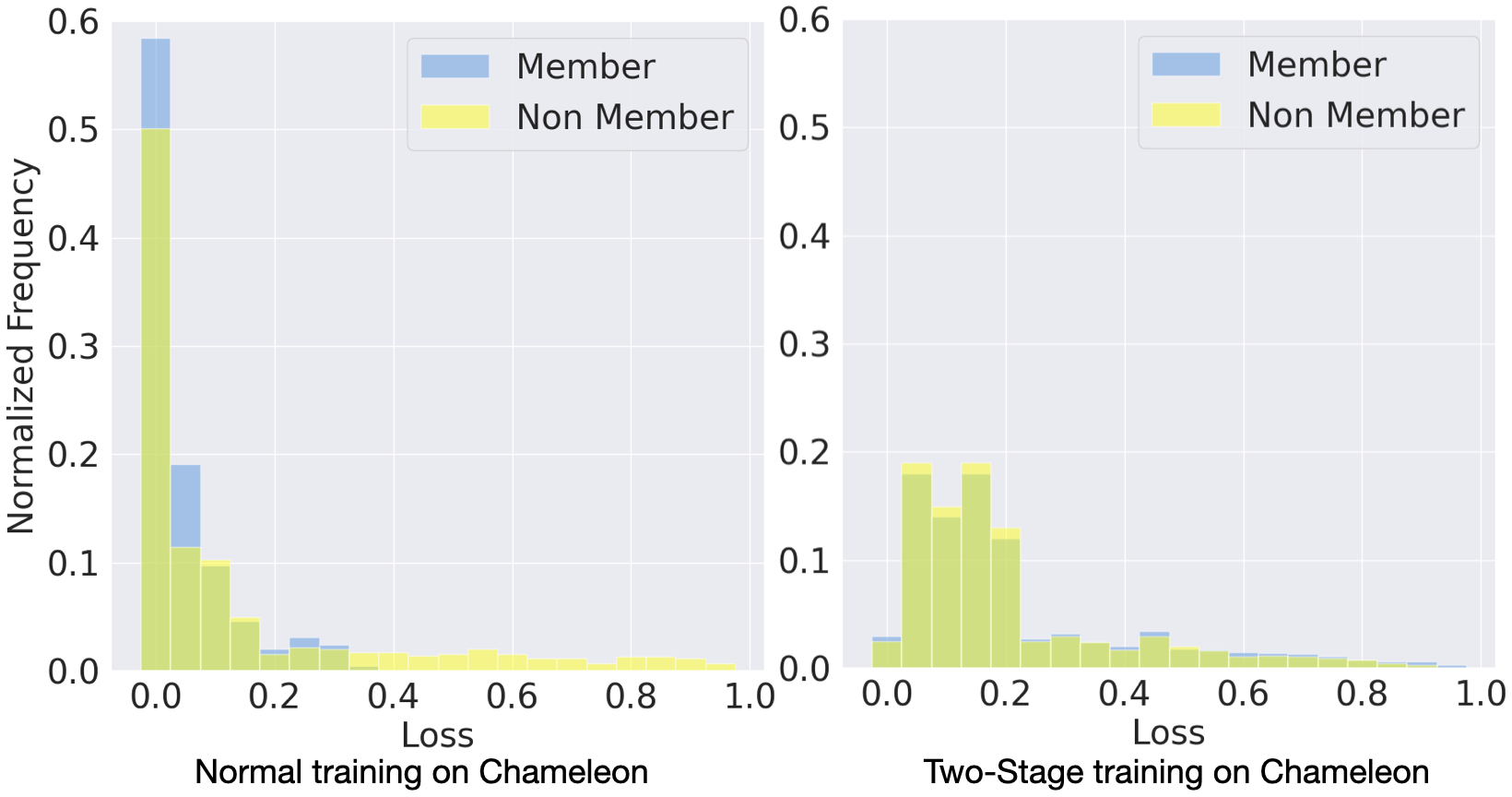}
        \caption{}
    \end{subfigure}
    \vspace{-0.05in}
    \caption{(a) The train and test loss of GCN on Cora dataset, with normal ($200$ epochs) and two-stage training ($400$ epochs) schedule. The difference between train and test loss after two-stage training is significantly smaller than normal training. (b) A diagram of the two-stage training schedule. The key idea is to use the predicted labels of test nodes from the first stage (blue) as psudolabels and switch the training and testing set in the second stage (yellow). More details are included in Section~\ref{sec:method}. (c)(d) Sample-wise loss distribution for train (member) and test (non-member) nodes with normal and two-stage training schedule on Cora and Chameleon dataset, respectively. The divergence between train and test loss distribution for two-stage training schedule is smaller than normal training one.}
    \label{fig:intro}
\vspace{-0.28in}
\end{figure*}

While recent studies have examined MIA for node classification tasks~\citep{olatunji2021membership,wu2021adapting,he2021node,conti2022label}, most focus on the inductive setting where training and test datasets have disjoint graph topologies. For example, the TSTS (train on subgraph, test on subgraph) approach in~\citep{olatunji2021membership} assumes no overlap between training and test subgraphs, effectively reducing the analysis to a standard graphless MIA problem. In contrast, MIA in the graph transductive setting remains underexplored, despite the prevalence of transductive graph learning in real-world applications. The challenge is three-fold. \textbf{(C1)} The definition of "membership" differs, as transductive GNNs have access to node features and neighbors of test nodes during training, with only label membership being unknown, while classical MIA assumes no knowledge about testset. \textbf{(C2)} The constant node features and graph topology across training and testing can intensify the difficulty of protecting label membership, as GNNs can be more prone to overfitting the \emph{combined} node and neighborhood representations compared to graphless models~\citep{bechler2023graph}, potentially degrading the performance of previous defense methods~\citep{olatunji2021membership,shejwalkar2020membership} when applied in the transductive setting, as shown in Section~\ref{subsec:two_stage_compare}. \textbf{(C3)} Evaluating transductive graph MIA requires a new framework, as simply splitting the original dataset into disjoint target and shadow datasets would violate the transductive learning assumption.

We take the initial step here to address these challenges, including a formal problem formulation of graph transductive MIA (C1), a simple yet effective two-stage defense mechanism, graph transductive defense (GTD) (C2), and a worst-case based analysis framework to ensure fair evaluation by reducing the splitting and noise randomness (C3). We begin by confirming that overfitting is one of the main contributors to GNN vulnerability to membership inference attacks in the transductive setting, as evidenced by the significantly lower train loss compared to testing loss at all training steps, as illustrated in Figure~\ref{fig:intro}(a). Consequently, we introduce an effective and specialized defense method tailored to transductive setting, which is depicted in Figure~\ref{fig:intro}(b). To counteract the overfitting effect, we adopt a flattening strategy~\citep{chen2022relaxloss} to increase the variance of the train loss distribution. Furthermore, drawing inspiration from graph self-supervised learning~\citep{liu2023graph}, we leverage the availability of the entire graph topology along with node features during training process to propose a two-stage, train-test alternate training procedure to further close the gap between the training and testing loss distributions, as depicted in Figure~\ref{fig:intro}(c) and (d). Notably, GTD allows for a utility-preserving (or even improving) defense compared to other perturbation-based defense approaches, and it can be seamlessly integrated into any classification model with minimal overhead. Our extensive empirical studies on both synthetic (contextual stochastic block model~\citep{deshpande2018contextual}) and nine real-world graph datasets demonstrate the superior performance of our proposed approach, compared against state-of-the-art defense methods for graphs and graphless models (a decrease in attack AUROC by $9.42\%, 4.98\%$ and an increase in utility performance by $18.08\%, 5.82\%$ on average compared to LBP~\citep{olatunji2021membership} and DMP~\citep{shejwalkar2020membership}, respectively). Lastly, we analyze the relationship between defense performance and graph topology, as well as dataset properties, which contributes to a better understanding of graph MIA within the graph learning community, providing valuable insights for future research.

\vspace{-0.05in}
\section{Related Works}
\vspace{-0.08in}
\label{sec:related_works}
Due to limited space, we provide only a brief summary of related works in the main text, with a more detailed description of each attack and defense method available in Appendix~\ref{app:related}.

\textbf{Membership Inference Attacks.} MIA on ML models aim to infer whether a data record was used to train a target ML model or not.  This concept is firstly proposed by~\citet{homer2008resolving} and later on extended to various directions, ranging from white-box setting~\citep{Nasr_2019,rezaei2021difficulty,melis2019exploiting,leino2020stolen}, to black-box setting~\citep{shokri2017membership,salem2018mlleaks,song2020systematic,li2021membership,choquette2021label,carlini2022membership}. Upon identifying the informative features (e.g., posterior predictions, loss values, gradient norms, etc.) that distinguish the sample membership, the attacker can choose to learn either a binary classifier~\citep{shokri2017membership} or metric-based decisions~\citep{yeom2018privacy,salem2018mlleaks} from shadow model trained on shadow dataset to extract patterns in these features among the training samples for identifying membership. A standard MIA process is included in Appendix~\ref{app:mia_process}.

\textbf{Defense Against Membership Inference Attacks.} As MIA exploit the behavioral differences of the target model on trainset and testset, most defense mechanisms work towards suppressing the common patterns that an optimal attack relies on. Popular defense methods include confidence score masking~\citep{shokri2017membership,jia2019memguard,yang2020defending,li2021mem,choquette2021label,hanzlik2021mlcapsule}, regularization~\citep{hayes2017logan,salem2018mlleaks,leino2020stolen,wang2020against,choquette2021label,kaya2021does,yu2021does,chen2022relaxloss}, knowledge distillation~\citep{shejwalkar2020membership,tang2022mitigating}, and differential privacy~\citep{naseri2020toward,saeidian2021quantifying}.

\textbf{Membership Inference Attacks and Defenses for GNNs.} There are a handful of research that focuses on extending MIA and corresponding defense mechanisms to graph learning framework.~\citet{olatunji2021membership} analyzed graph MIA in two settings (train on subgraph, test on subgraph/full), and proposed the LBP defense based on the confidence score masking idea,~\citet{he2021node} proposed zero-hop and two-hop attacks designed for inductive GNNs,~\citet{wang2023link} studied the link membership inference problem in an unsupervised fashion, and~\citet{chen2024maskarmor} developed MaskArmor based on masking and distillation technique. Nevertheless, no existing work has explored the intersection of MIA and graph transductive learning setting (i.e., node classification task in a supervised manner), and this paper aims to fill this gap in between.

\vspace{-0.1in}
\section{Formulation of MIA in Graph Transductive Setting}
\vspace{-0.08in}
\label{sec:formulation}
In this paper we focus on the supervised node classification tasks in transductive setting; nevertheless, our method is applicable to different graph learning scenarios. Let $\mathcal{G} = (X, A, Y, \mathcal{V}^{\text{Train}}, \mathcal{V}^{\text{Test}})$ denote the graph dataset with node features $X\in\mathbb{R}^{n\times d}$, adjacent matrix $A\in\mathbb{R}^{n\times n}$, one-hot encoded node labels $Y\in\mathbb{R}^{n\times C}$, trainset $\mathcal{V}^{\text{Train}}$ and testset $\mathcal{V}^{\text{Train}}$. Here $n$ is the number of nodes, $d$ is the feature dimension, $C$ is the number of classes, $\mathcal{V}^{\text{Train}}$ and $\mathcal{V}^{\text{Test}}$ are disjoint and $|\mathcal{V}^{\text{Train}}| + |\mathcal{V}^{\text{Test}}|=n$. We later on use $Y^\text{Train}$ to denote the labels of $\mathcal{V}^{\text{Train}}$, and $\hat{Y}^\text{Test}$ to denote the predicted labels of $\mathcal{V}^{\text{Test}}$. Since $X$ and $A$ are already known during training, the goal of \textbf{graph transductive MIA} is determing the \textbf{label membership}: given a node $v\in\mathcal{V}^{\text{Train}}\cup\mathcal{V}^{\text{Test}}$, determine if $v \in \mathcal{V}^{\text{Train}}_{t}$ (member) or not (non-member). Following the practice of MIA, we also need a shadow dataset $\mathcal{G}_s = (X_s, A_s, Y_s, \mathcal{V}^{\text{Train}}_s, \mathcal{V}^{\text{Test}}_s)$ to train the shadow model, and the choice of $\mathcal{G}_s$ is explained in more details in Section~\ref{sec:worst_case}.


It is worth pointing out the difference between inductive and transductive MIA: in inductive setting, the testset is not used by the target model during training, making graph inductive MIA quite similar to graphless MIA. In contrast, the transductive setting exposes and incorporates part of the testset information (such as node features and neighborhoods) during training. Consequently, transductive GNNs can learn to differentiate the topological characteristics between the trainset (considered in the loss) and the testset (not considered in the loss), complicating the protection of label membership. Supporting this, we empirically demonstrate in Section~\ref{subsec:topology_analysis} that incorporating more topological information can in general enhance the attack performance.

\vspace{-0.12in}
\section{Two-Stage Defense Method Graph Transductive Defense}
\vspace{-0.05in}
\label{sec:method}

\begin{wrapfigure}[34]{L}{0.5\textwidth}
\begin{minipage}{0.5\textwidth}
\vspace{-24pt}
\begin{algorithm}[H]
    \caption{GTD Training Procedure}
    \label{alg:gtd}
    \begin{algorithmic}
        \STATE \textbf{Input:} Dataset $\mathcal{G} = (X, A, Y, \mathcal{V}^{\text{Train}}, \mathcal{V}^{\text{Test}})$, training epochs $E$, learning rates $\gamma$,  number of classes $C$, loss threshold $\alpha$, flattening param $\beta$
        \STATE \textbf{Output:} Second stage target model $M_{2}({\theta_{2}})$
        \STATE \textbf{First stage:}
        \STATE Initialize first stage model $M_{1}({\theta_{1}})$ with random initializations
        \FOR{epoch in $[1,E]$}
            \IF{loss $L(M_{1}(\theta_{1}), Y^{\text{Train}}) \ge \alpha$}
            \STATE $\theta_{1} \gets \theta_{1} - \gamma \cdot \nabla_{\theta_1} L(M_{1}(\theta_{1}), Y^{\text{Train}})$
            \ELSE
                \STATE Construct soft labels
                $S^{\text{Train}}$ where $s_c^{\text{Train}}=
                \begin{cases}
                 \beta, \text { if } y_{c}^{\text{Train}}=1; \\ 
                 (1-\beta)/(C-1), \text { otherwise }
                \end{cases}$
                \STATE $\theta_{1} \gets \theta_{1} - \gamma \cdot \nabla_{\theta_1} L(M_{1}(\theta_{1}), S^{\text{Train}})$
            \ENDIF
        \ENDFOR
        \STATE \textbf{Second stage:}
        \STATE Initialize second stage model $M_{2}({\theta_{2}})$ with $M_{1}({\theta_{1}})$, and generate psudolabels $\hat{Y}^{\text{Test}}$ for original testset by inferencing $M_{1}({\theta_{1}})$
        \FOR{epoch in $[1,E]$}
            \IF{loss $L(M_{2}(\theta_{2}), \hat{Y}^{\text{Test}}) \ge \alpha$}
            \STATE $\theta_{2} \gets \theta_{2} - \gamma \cdot \nabla_{\theta_2} L(M_{2}(\theta_{2}), \hat{Y}^{\text{Test}})$
            \ELSE
                \STATE Construct soft labels
                $S^{\text{Test}}$ where $s_c^{\text{Test}}=
                \begin{cases}
                 \beta, \text { if } \hat{y}_{c}^{\text{Test}}=1; \\ 
                 (1-\beta)/(C-1), \text { otherwise }
                \end{cases}$
                \STATE $\theta_{2} \gets \theta_{2} - \gamma \cdot \nabla_{\theta_2} L(M_{2}(\theta_{2}), S^{\text{Test}})$
            \ENDIF
        \ENDFOR
    \end{algorithmic}
\end{algorithm}
\end{minipage}
\end{wrapfigure}

To defense MIA in graph transductive setting, we introduce a \textbf{two-stage} defense GTD, depicted in Algorithm~\ref{alg:gtd}, to train \textbf{target GNNs}. The motivation of the method is to reduce the gap between training and testing loss distributions and alleviate the \textbf{overfitting} of target models on trainset.

The first stage of GTD involves using $\mathcal{G}$ to train a model checkpoint $M_{1}({\theta_{1}})$ with parameters denoted as $\theta_1$. We adopt the flattening strategy in the first stage as a regularization, inspired by~\citep{chen2022relaxloss}. The flattening is implemented by transforming hard labels (one-hot) to
soft labels (probability vector) when the loss on trainset falls below the threshold $\alpha$. For simplicity, we assign the value $\beta$ to the groundtruth class, and $\frac{1-\beta}{C-1}$ to the others. Here $\alpha,\beta$ are two hyperparameters. Note that we only use soft labels to compute loss when the loss is small enough to keep the model utility as high as possible. The key of flattening is to increase the mean and variance of training loss distribution, as we are introducing noise to label distribution. By flattening, training loss distribution can have larger overlap with testing one, making it harder for attackers to implement MIA.


The second stage of GTD is similar to the first one, with the main difference that we instead use $(X, A, \hat{Y}^{\text{Test}})$ for training. The psudolabels $\hat{Y}^{\text{Test}}$ are generated by inferencing the checkpoint $M_{1}({\theta_{1}})$ on testset. Instead of random initialization, we also initialize the second stage model $M_{2}({\theta_{2}})$ with the checkpoint to resume training. The subsequent training process also proceeds with flattening, and $M_{2}({\theta_{2}})$ is the final output target model. The gist of the second stage is to also involve testset into training, even when we do not have access to their groundtruth labels $Y^{\text{Test}}$. In this case, the testset is also ``trained'', as they go through the same procedure as trainset.


Compared to state-of-the-art defense methods based on perturbations and distillation, such as LBP~\citep{olatunji2021membership} for GNNs and DMP~\citep{shejwalkar2020membership} for graphless models, GTD can achieve a better balance between model utility and defense performance. LBP employs noise addition to the posteriors of the target model, grouping the elements randomly and adding noise from the same Laplace distribution to each group to reduce the required amount of noise. While LBP offers strong defense capabilities, the added noise significantly degrades the target model utility. DMP, on the other hand, tunes the data used for knowledge transfer to enhance membership privacy. It utilizes an unprotected model trained on private data to guide the training of a protected target model on reference data, optimizing the tradeoff between membership privacy and utility. However, DMP necessitates the collection of an additional dataset for training the protected model, which complicates the whole process. Meanwhile, our method addresses the overfitting problem implicitly by ensuring that both the training set and test set undergo the same procedure. Consequently, our method offers several advantages over LBP and DMP: (1) it avoids explicitly adding noise to the target model predictions, thereby preserving model utility; (2) it does not require additional data; and (3) it fully leverages the characteristics of the graph transductive setting through a train-test alternate training schedule.

\vspace{-0.1in}
\section{Experiments}
\vspace{-0.05in}
\label{sec:exp}
\textbf{Datasets and GNN Baselines.}
We train four GNN (GCN~\citep{kipf2016semi}, GAT~\citep{velickovic2017graph}, SGC~\citep{wu2019simplifying}, GPRGNN~\citep{chien2020adaptive}) on six homophilic datasets (Cora, CiteSeer, PubMed~\citep{https://doi.org/10.1609/aimag.v29i3.2157, pmlr-v48-yanga16}, Computers, Photo~\citep{10.1145/2766462.2767755, shchur2019pitfalls}, Ogbn-Arxiv~\citep{NEURIPS2020_fb60d411}), and four GNN (NLGCN, NLGAT, NLMLP~\citep{wang2018non}, GPRGNN) on three heterophilic datasets (Texas, Chameleon, Squirrel~\citep{10.1093/comnet/cnab014}). Detailed experiment setup, properties and statistics of datasets are relegated to Appendix~\ref{app:complete_exp}.

\textbf{Evaluation Metrics.} We use two metrics for evaluation. We report classification accuracy of the target model on testset to measure model utility, and AUROC scores of the attack model, which is widely used in the field of MIA~\citep{carlini2022membership}, to fairly measure the defense capability.


We summarize the main research questions that we try to investigate in this section: 
\textbf{(RQ1)} Can GTD outperform other start-of-the-art defense approaches? 
\textbf{(RQ2)} What changes does GTD bring to the target model?
\textbf{(RQ3)} Which component of GTD contributes most to the performance improvement?
\textbf{(RQ4)} Will different graph topologies affect GTD defense capabilities?
\vspace{-0.05in}
\subsection{Worst-Case Analysis Framework}
\vspace{-0.02in}
\label{sec:worst_case}
For simplicity, we assume that the shadow dataset $\mathcal{G}_{s} = (X, A, Y_{s}^{\text{Train}}, \mathcal{V}^{\text{Train}}_{s}, \mathcal{V}^{\text{Test}}_{s})$ shares the same underlying features and graph topology with the target dataset. In this case, the worst-case for the target model (the best-case for the attacker) is that the shadow trainset and labels match exactly with the target trainset. Consequently, the trained shadow model's functionality is maximally similar to that of the target model, resulting in an optimal attacker theoretically. We denote this as the \textbf{hard} setting, and all our following results are obtained in the hard setting unless specified.


It is important to clarify that training an attack model in the hard setting does not imply that the attacker has complete knowledge of the target dataset's label membership information; if that were the case, the MIA would be trivial. We adopt the hard setting mainly for evaluation purposes because it allows us to: 1) establish a lower bound on the performance of different defense methods, and 2) minimize excessive variance in experimental results caused by the randomness in sampling the shadow training set and shadow training labels.


\vspace{-0.05in}
\subsection{Comparison with Other Defenses}
\vspace{-0.02in}
\label{subsec:two_stage_compare}
To answer \textbf{RQ1}, we choose two representative defense methods, Laplacian Binned Posterior Perturbation (LBP) on GNNs and Distillation for Membership Privacy (DMP) on graphless models, as our defense baselines. LBP is the state-of-the-art defense method for GNNs by adding Laplacian noise to the posterior before it is released to the user. To reduce the amount of noise needed to distort the posteriors, LBP doesn't add noise to each element of the posterior, but to binned posterior. In our experiments, we first randomly shuffle the posteriors and then assign each posterior to a partition/bin. The total number of bins $N$ is predefined based on the number of classes. For each bin, we sample noise at scale $b$ from the Laplace distribution. The sampled noise is added to each element in the bin. After the noise added to each bin, we restore the initial positions of the noisy posteriors and release them. Appendix Table~\ref{tab:LBP Parameters} shows the best set of parameters for LBP that we used in our experiments.

On the other hand, we adapted DMP to the case of GNN training. DMP consists of three phases, namely pre-distillation, distillation and post-distillation. The pre-distillation phase trains an unprotected model on a private training data without any privacy protection. Next, in the distillation phase, DMP selects reference data and transfers the knowledge of the unprotected model into predictions of the reference data. Notice that private training data and reference data have \textbf{no} intersection. Finally, In the post-distillation phase, DMP uses the predictions to train a protected model. Our experiments used the same model structure for the unprotected and protected models. To follow the procedure of DMP, we need to further split the trainset into private datasets and reference datasets, where the private datasets trains the unprotected models, and the reference datasets trains the protected target model. Compared to DMP, GTD can directly train target model with the full trainset.

In our experiments, the split ratio of trainsets and testsets for GTD and LBP is 1:1, and the split ratio of private datasets, reference datasets and testsets in DMP is 0.45:0.45:0.1. Table~\ref{tab:LBP small} and Table~\ref{tab:DMP small} shows partial result of our experiments, and the complete results can be found in Appendix~\ref{Comparison with LBP} and~\ref{Comparison with DMP}. In both tables, "Classify Acc" measures the utility performance of target models on testset after applying the defense methods and "Attack AUROC" shows the AUROC of attack models. Notice that better defense method should have higher "Classify Acc" and lower "Attack AUROC".
The results indicate that our method achieves better performance in both model utility and defense capability on all datasets and GNN backbones, compared to LBP and DMP. Specifically, compared to LBP, we significantly improved the model utility by 12.68\% while achieving higher defense capabilities  by 44.81\% on Chameleon with NLMLP. The main reason is that LBP is a perturbation-based method, which can potentially hurt the target model performance significantly. However, our method achieves defense by alleviating overfitting, which delves deeper into the core issue, instead of adversely affecting target models. In addition, it is also worth pointing that, compared to DMP, we achieve more pronounced improvement on small datasets (i.e., Cora, CiteSeer) and simpler model architectures (i.e., SGC, NLMLP) . For example, we significantly improved the model utility by 6.68\% and defense capabilities by 24.55\% on Chameleon with NLMLP. This gain is expected, as our method not only can make use of the full trainsets, but also utilizes testsets in the second stage, thus enhancing the model's generalization ability. 

\begin{table*}
   \caption{Performance comparison between GTD and LBP. Compared to LBP, GTD achieves a decrease in attack AUROC by $9.42\%$ and an increase in utility performance by $18.08\%$ on average.}
   \label{tab:LBP small}
   \centering
   \resizebox{\textwidth}{!}{
   \begin{tabular}{llllll}
    \toprule
    Dataset & Models & Classify Acc (LBP) &Classify Acc (GTD) &Attack AUROC (LBP) &Attack AUROC (GTD)\\
    \midrule
    \multirow{4}{*}{Cora}
    & GCN& 0.8321 $\pm$ 0.0052& 0.8684 $\pm$ 0.0031 & 0.5115 $\pm$ 0.0043& 0.4659 $\pm$ 0.0042\\
    & GAT & 0.8341 $\pm$ 0.0043& 0.8657 $\pm$ 0.0018 & 0.5196 $\pm$ 0.0057& 0.4522 $\pm$ 0.0056\\
    & SGC & 0.8321 $\pm$ 0.0065& 0.8731 $\pm$ 0.0054 & 0.5216 $\pm$ 0.0072& 0.4977 $\pm$ 0.0068\\
    & GPRGNN& 0.8601 $\pm$ 0.0050& 0.8806 $\pm$ 0.0030 & 0.5287 $\pm$ 0.0060& 0.4894 $\pm$ 0.0058\\
    \midrule
    \multirow{4}{*}{CiteSeer} 
    & GCN& 0.6980 $\pm$ 0.0047 & 0.7233 $\pm$ 0.0035 & 0.5138 $\pm$ 0.0060 & 0.4035 $\pm$ 0.0061\\
    & GAT& 0.6952 $\pm$ 0.0051 & 0.7210 $\pm$ 0.0042 & 0.5190 $\pm$ 0.0047 & 0.3789 $\pm$ 0.0049\\
    & SGC& 0.7013 $\pm$ 0.0074 & 0.7362 $\pm$ 0.0065 & 0.5268 $\pm$ 0.0081 & 0.4499 $\pm$ 0.0069\\
    & GPRGNN& 0.7210 $\pm$ 0.0058& 0.7358 $\pm$ 0.0047 & 0.5288 $\pm$ 0.0065& 0.4619 $\pm$ 0.0062\\
    \midrule
    \multirow{4}{*}{PubMed} 
    & GCN& 0.6886 $\pm$ 0.0041 & 0.8381 $\pm$ 0.0023 & 0.4998 $\pm$ 0.0050 & 0.4990 $\pm$ 0.0048\\
    & GAT & 0.7631 $\pm$ 0.0037 & 0.8400 $\pm$ 0.0028 & 0.5021 $\pm$ 0.0084 & 0.4911 $\pm$ 0.0061\\
    & SGC & 0.6564 $\pm$ 0.0035 & 0.8080 $\pm$ 0.0020 & 0.5007 $\pm$ 0.0065 & 0.5005 $\pm$ 0.0057\\
    & GPRGNN& 0.7843 $\pm$ 0.0029 & 0.8553 $\pm$ 0.0014 & 0.5003 $\pm$ 0.0038 & 0.4967 $\pm$ 0.0034\\
    \midrule
    \multirow{4}{*}{Chameleon} 
    & NLGCN& 0.5987 $\pm$ 0.0068&0.6657 $\pm$ 0.0062&0.5033 $\pm$ 0.0062&0.4854 $\pm$ 0.0065\\
    & NLGAT& 0.5926 $\pm$ 0.0072&0.6585 $\pm$ 0.0070&0.5046 $\pm$ 0.0065&0.4602 $\pm$ 0.0063\\
    & NLMLP& 0.4281 $\pm$ 0.0078&0.4824 $\pm$ 0.0074&0.5523 $\pm$ 0.0057&0.3048 $\pm$ 0.0051\\
    & GPRGNN& 0.5230 $\pm$ 0.0054&0.6550 $\pm$ 0.0058&0.5107 $\pm$ 0.0060&0.4936 $\pm$ 0.0049\\
    \bottomrule
    \end{tabular}}
\end{table*}

\begin{table*}[htbp]
  \caption{Performance comparison between GTD and DMP. Compared to DMP, GTD achieves a decrease in attack AUROC by $4.98\%$ and an increase in utility performance by $5.82\%$ on average.}
  \centering
  \label{tab:DMP small}
  \resizebox{\textwidth}{!}{
  \begin{tabular}{llllll}
    \toprule
    Dataset & Models & Classify Acc (DMP) & Classify Acc (GTD) & Attack AUROC (DMP) & Attack AUROC (GTD) \\
    \midrule
    \multirow{4}{*}{Cora}
     & GCN & 0.7646 $\pm$ 0.0058 & 0.8842 $\pm$ 0.0058 & 0.5136 $\pm$ 0.0051& 0.5043 $\pm$ 0.0060 \\
     & GAT & 0.7452 $\pm$ 0.0050 & 0.8858 $\pm$ 0.0047 & 0.5104 $\pm$ 0.0053& 0.5005 $\pm$ 0.0049 \\
     & SGC & 0.7521 $\pm$ 0.0071& 0.8902 $\pm$ 0.0041 & 0.5096 $\pm$ 0.0046& 0.4994 $\pm$ 0.0051 \\
     & GPRGNN & 0.6323 $\pm$ 0.0048& 0.8926 $\pm$ 0.0024 & 0.5162 $\pm$ 0.0040& 0.5044 $\pm$ 0.0043 \\
    \midrule
    \multirow{4}{*}{CiteSeer}
     & GCN & 0.7434 $\pm$ 0.0058& 0.7628 $\pm$ 0.0062 & 0.5202 $\pm$ 0.0053& 0.4811 $\pm$ 0.0045 \\
     & GAT & 0.7156 $\pm$ 0.0061& 0.7564 $\pm$ 0.0059 & 0.5176 $\pm$ 0.0059& 0.4766 $\pm$ 0.0045 \\
     & SGC & 0.7403 $\pm$ 0.0068& 0.7692 $\pm$ 0.0049 & 0.5178 $\pm$ 0.0050& 0.4814 $\pm$ 0.0041 \\
     & GPRGNN & 0.7426 $\pm$ 0.0039& 0.7726 $\pm$ 0.0028 & 0.5204 $\pm$ 0.0045& 0.4962 $\pm$ 0.0038 \\
    \midrule
    \multirow{4}{*}{PubMed}
     & GCN & 0.8235 $\pm$ 0.0037& 0.8387 $\pm$ 0.0034 & 0.5026 $\pm$ 0.0039& 0.4978 $\pm$ 0.0042 \\
     & GAT & 0.8027 $\pm$ 0.0047& 0.8434 $\pm$ 0.0024 & 0.5013 $\pm$ 0.0036& 0.5005 $\pm$ 0.0043 \\
     & SGC & 0.8013 $\pm$ 0.0041& 0.8096 $\pm$ 0.0045 & 0.5024 $\pm$ 0.0042& 0.5003 $\pm$ 0.0038 \\
     & GPRGNN & 0.8104 $\pm$ 0.0031& 0.8423 $\pm$ 0.0036 & 0.5020 $\pm$ 0.0027& 0.4994 $\pm$ 0.0023 \\
    \midrule
    \multirow{4}{*}{Chameleon}
     & NLGCN & 0.6681 $\pm$ 0.0064& 0.6963 $\pm$ 0.0065 & 0.5210 $\pm$ 0.0064& 0.5182 $\pm$ 0.0062 \\
     & NLGAT & 0.6516 $\pm$ 0.0066& 0.7082 $\pm$ 0.0073 & 0.5116 $\pm$ 0.0061& 0.5159 $\pm$ 0.0059 \\
     & NLMLP & 0.4643 $\pm$ 0.0079& 0.4955 $\pm$ 0.0070 & 0.5054 $\pm$ 0.0053& 0.3817 $\pm$ 0.0056 \\
     & GPRGNN & 0.6471 $\pm$ 0.0060& 0.6934 $\pm$ 0.0068 & 0.5172 $\pm$ 0.0067& 0.5163 $\pm$ 0.0045\\
    \bottomrule
  \end{tabular}}
\end{table*}

\subsection{Generalization Gap after Two-Stage Training}

\begin{figure*}[htbp]
    \centering
    \includegraphics[width=\linewidth]{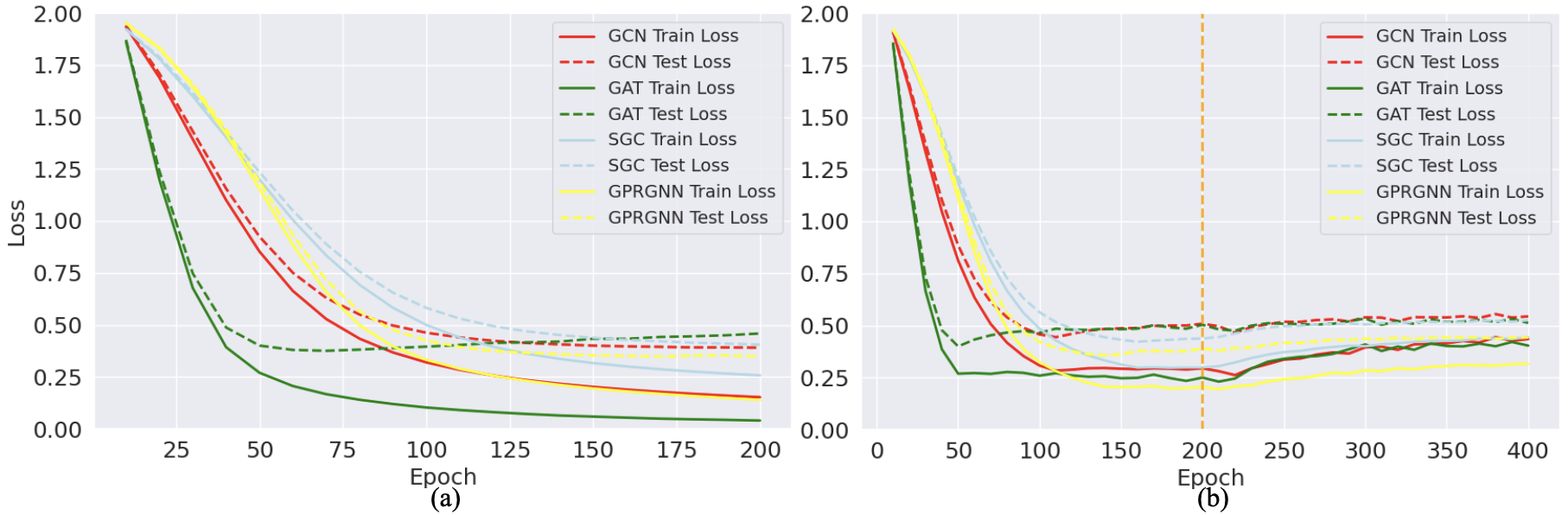}
    \caption{Comparison of average training loss and testing loss on Cora for (a) normal training, (b) two-stage training (GTD). In (b), the left half of the orange dashed line indicates the first training stage of our method, while the one on the right indicates the second stage.}
    \label{fig:Cora_Train_Test_Loss}
    \vspace{-10pt}
\end{figure*}

In this section, we analyzed the changes in training loss and testing loss distribution before and after GTD training (\textbf{RQ2}). Experiments are conducted on the homophilic dataset Cora and the heterophilic dataset Chameleon, with GCN and NLGCN, respectively.

Through experiments, we demonstrated that GTD can: (1) reduce the gap between the average losses of training and testing nodes, thereby alleviating overfitting; (2) increase the variance of both member and non-member loss distributions and reduce the disparity between their means; and (3) decrease the distinguishability between member and non-member loss distributions.

\textbf{Reduce the gap between the average losses of training and testing nodes.} Figure~\ref{fig:Cora_Train_Test_Loss} shows the variations of the average losses of training and testing nodes with increasing training epochs for both normal training and two-stage training on Cora dataset. The result on Chameleon dataset is shown in Appendix Figure~\ref{fig:Chameleon_Train_Test_Loss}. We also recorded the losses of all models from Figure~\ref{fig:Cora_Train_Test_Loss} and Figure~\ref{fig:Chameleon_Train_Test_Loss} at the end of training to Appendix Table~\ref{tab:normal vs gtd}, and additionally added the result of comparative experiments on model utility and defense capability. Comparing Figure~\ref{fig:Cora_Train_Test_Loss} (a) with (b), it can be observed that the difference between the average losses of training and testing nodes in the normal training increases as epochs increase, indicating that overfitting exists and becomes worse as training proceeds. However, when using two-stage training, although overfitting cannot be completely avoided, the difference between training and testing losses decreases in the second stage as training proceeds, indicating a gradual alleviation of overfitting. Table~\ref{tab:normal vs gtd} also shows that our method achieved lower average loss gap after the entire training process. All experimental results demonstrate the capability of our method to reduce overfitting and the generalization gap.

\begin{figure*}[htbp]
    \centering
    \includegraphics[width=\linewidth]{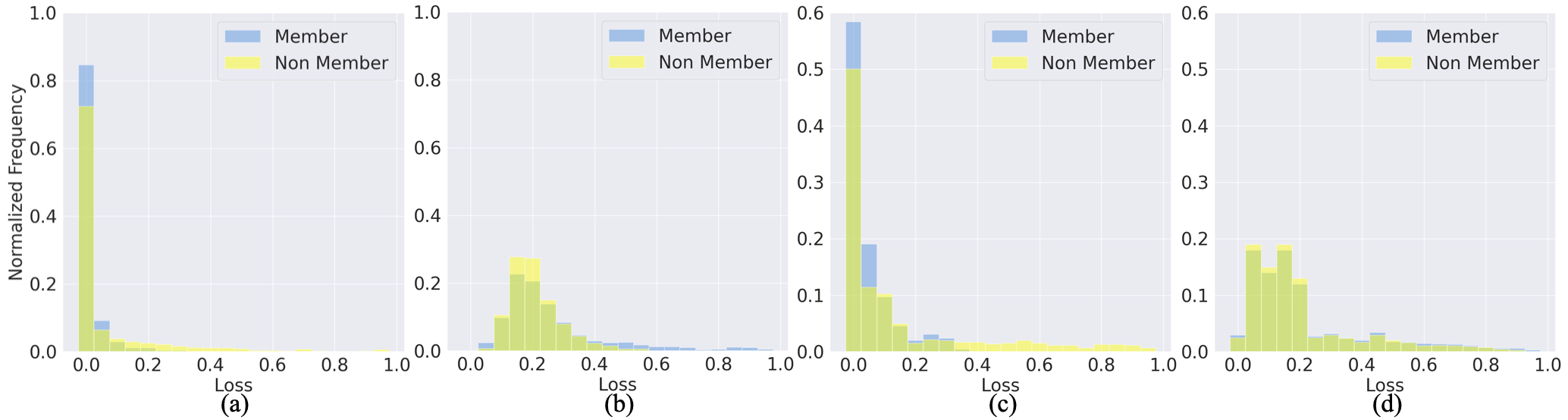}
    \caption{Loss distribution histograms for (a) normal training on Cora, (b) two-stage training (GTD) on Cora, (c) normal training on Chameleon, and (d) two-stage training (GTD) on Chameleon.}
    \label{fig:Train_Test_Loss_Distribution}
    \vspace{-10pt}
\end{figure*}

\textbf{Increase the variance of both member and non-member loss distributions and reduce the disparity between their means.} Figure~\ref{fig:Train_Test_Loss_Distribution} illustrates the loss distributions of member and non-member nodes on the Cora and Chameleon datasets after training with both normal and two-stage methods. According to the definition in Section~\ref{sec:formulation}, members refer to the nodes in the trainset of the target model, while non-members refer to the nodes in the testset. Therefore, Figure~\ref{fig:Train_Test_Loss_Distribution} can also be viewed as the training and testing loss distributions of the target model after using different training methods. Comparing Figure~\ref{fig:Train_Test_Loss_Distribution} (a) with (b) and (c) with (d), it can be observed that the loss distributions of members and non-members after normal training have relatively small variances, and their means differ significantly. This conclusion is consistent with the results of the average losses in Table~\ref{tab:normal vs gtd}. However, after using two-stage training, significant changes occur in the loss distributions: the variances of both two distributions increase significantly. And combined with the results in Table~\ref{tab:normal vs gtd}, it is obvious that their means become closer.

\textbf{Decrease the distinguishability between member and non-member loss distributions.} From Figure~\ref{fig:Train_Test_Loss_Distribution}, it can be seen that the overlap between the member and non-member loss distributions of the target model after two-stage training is significantly larger than that of normal training. Combined with the conclusions obtained above, we can confirm that the distinguishability between member and non-member distributions has decreased, which will increase the difficulty of MIA. 

In summary, the changes of the target model induced by our two-stage training method are significant. Table~\ref{tab:normal vs gtd} also demonstrates that such changes not only substantially enhance defense capability but also result in only subtle decline in downstream classification accuracy.
\vspace{-5pt}
\subsection{Ablation Study}
\begin{table*}[htbp]
\vspace{-5pt}
  \caption{Ablation study on the source of gains for GTD.}
  \label{tab:Ablation}
  \resizebox{\textwidth}{!}{
  \begin{tabular}{lllll}
    \toprule
    Method & Dataset & GNN Models & Classify Acc &Attack AUROC \\
    \midrule    
    \multirow{4}{*}{Normal Training} & Cora
    & GCN & 0.8742 $\pm$ 0.0021& 0.5712 $\pm$ 0.0035\\
    & CiteSeer
    & GCN& 0.7296 $\pm$ 0.0034& 0.5979 $\pm$ 0.0048 \\
    & PubMed
    & GCN & 0.8423 $\pm$ 0.0027& 0.5206 $\pm$ 0.0032\\
    & Chameleon
    & NLGCN& 0.6817 $\pm$ 0.0054& 0.5469 $\pm$ 0.0051\\
    \midrule
    \multirow{4}{*}{Flattening (One-Stage)} 
    & Cora
    & GCN & 0.8674 $\pm$ 0.0028& 0.5617 $\pm$ 0.0045\\
    & CiteSeer
    & GCN& 0.7234 $\pm$ 0.0036& 0.5886 $\pm$ 0.0063 \\
    & PubMed
    & GCN & 0.8345 $\pm$ 0.0025& 0.5078 $\pm$ 0.0041\\
    & Chameleon
    & NLGCN& 0.6720 $\pm$ 0.0068& 0.5327 $\pm$ 0.0055\\
    \midrule    
    \multirow{4}{*}{Flattenning \& Gradient Asent (One-Stage) } & Cora
    & GCN & 0.8636 $\pm$ 0.0038& 0.5586 $\pm$ 0.0053\\
    & CiteSeer
    & GCN& 0.7221 $\pm$ 0.0046& 0.5857 $\pm$ 0.0070\\
    & PubMed
    & GCN & 0.8310 $\pm$ 0.0043& 0.5082 $\pm$ 0.0050\\
    & Chameleon
    & NLGCN& 0.6707 $\pm$ 0.0068& 0.5302 $\pm$ 0.0074 \\
    \midrule
    \multirow{4}{*}{Two-Stage (without Flattening)} 
    & Cora
    & GCN & 0.8706 $\pm$ 0.0030& 0.4834 $\pm$ 0.0045\\
    & CiteSeer
    & GCN& 0.7218 $\pm$ 0.0042& 0.4438 $\pm$ 0.0066\\
    & PubMed
    & GCN & 0.8328 $\pm$ 0.0035& 0.5039 $\pm$ 0.0043\\
    & Chameleon
    & NLGCN& 0.6745 $\pm$ 0.0060& 0.4977 $\pm$ 0.0063\\
    \midrule
    \multirow{4}{*}{GTD (Two-Stage \& Flattening)} & Cora
    & GCN& 0.8684 $\pm$ 0.0031& 0.4659 $\pm$ 0.0042\\
    & CiteSeer
    & GCN& 0.7233 $\pm$ 0.0035& 0.4035 $\pm$ 0.0061\\
    & PubMed
    & GCN& 0.8381 $\pm$ 0.0023& 0.4990 $\pm$ 0.0048\\
    & Chameleon
    & NLGCN& 0.6657 $\pm$ 0.0062& 0.4854 $\pm$ 0.0065\\
    \bottomrule
    \end{tabular}}
\end{table*}
Our two-stage defense method differs from conventional training methods in two aspects: (1) flattening operation and (2) two-stage training. To demonstrate their roles in enhancing defense capability, we conducted the following ablation experiments to answer \textbf{RQ3}.

In the experiments, we set up four variants: (1) normal training, (2) two-stage (without flattening), (3) flattening (one-stage), and (4) GTD. Here, normal training indicates training a target model only on the trainset; two-stage trains a target model in a train-test alternate fashion, equivalent to GTD without flattening; flattening is the same as described in Section~\ref{sec:method}, combined with one-stage training. Clearly, GTD is two-stage combined with flattening. In this set of comparison, we also considered RelaxLoss~\citep{chen2022relaxloss}, which is essentially a combination of alternate flattening and gradient ascent when the training loss falls below a predefined threshold. We use RelaxLoss as an example to show the difference between the defense methods effective for graph and graphless models, and necessities to design defense mechanisms specially for graph models.

Table~\ref{tab:Ablation} presents the results of ablation study regarding these four variants on four datasets and one GNN backbone GCN. The complete results can be found in Appendix~\ref{Complete Results of the Ablation Study}. 
The findings indicate that the primary source of improvement for GTD is the two-stage training technique. This method ensures that the testset undergoes the same process as the trainset, thus preserving the final model performance. Compared to flattening, the extra gradient ascent operation barely brings new gains in either model utility or defense capability in graph learning cases; meanwhile, gradient ascent is shown to be useful to defend non-graph MIA.

\vspace{-5pt}
\subsection{Influences of Graph Topology}
\label{subsec:topology_analysis}
To facilitate the analysis in this section, we also introduce a \textbf{weak} attack setting here to analyze the influence of graph topology. Compared to the hard setting, the weak counterpart refers to the scenario where the shadow trainset and the target trainset have minimal intersection. To be specific, we choose the shadow trainset $\mathcal{V}^{\text{Train}}_{s}=\text{argmin}_{\mathcal{V}}\left|\mathcal{V}\cap\mathcal{V}^{\text{Train}}\right|$ while keep the same trainset size, $|\mathcal{V}^{\text{Train}}_{s}|=|\mathcal{V}^{\text{Train}}|$.

\begin{wrapfigure}[17]{L}{0.59\textwidth}
\begin{minipage}{0.59\textwidth}
\vspace{-24pt}
\begin{figure}[H]
  \centering
  \includegraphics[width=\linewidth]{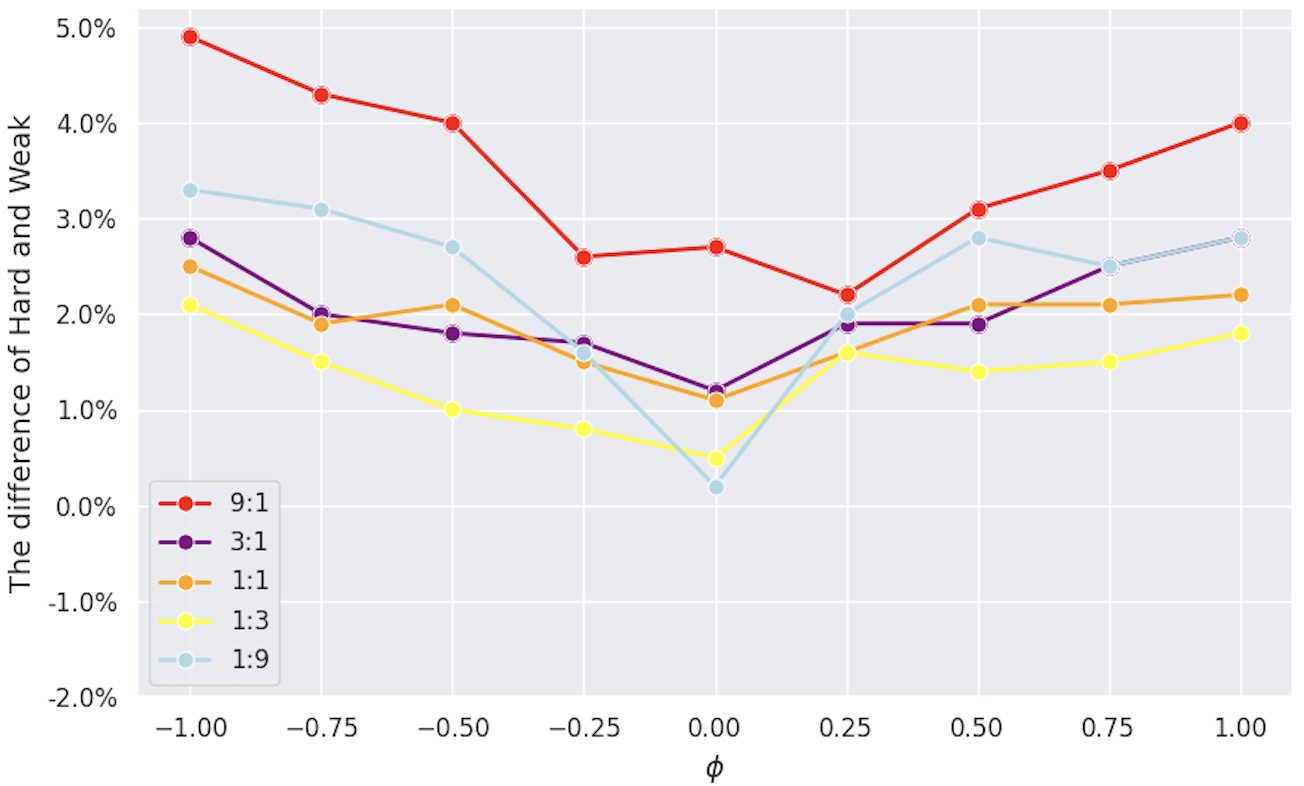}
  \caption{The difference of Attack AUROC between hard and weak setting (i.e., hard - weak). Different legends correspond to different train/test split ratios.}
  \label{fig:sythetic}
\end{figure}
\end{minipage}
\end{wrapfigure}

As the intensity of MIA is different in hard and weak settings, the defense capability of GTD is also different. However, we found that this difference is correlated with the graph topology. To investigate this correlation (\textbf{RQ4}), we conducted experiments on the cSBM synthetic dataset, for which we can change the level of homophily and heterophily by the hyperparameter $\phi$. The closer $\phi$ is to $1$, the more homophily the graph is; the closer $\phi$ is to $-1$, the more heterophily the graph is; when $\phi=0$, there is no graph information and the problem degrades to a graphless case. The detailed description about cSBM can be found in Appendix~\ref{cSBM}.

Appendix Table~\ref{tab:Synthetic1} shows the experiment result on cSBM synthetic datasets, and we plot the difference of attack AUROC under hard and weak settings in Figure~\ref{fig:sythetic}. At the same split ratio, when $\phi$ varies from $-1$ to $1$, the AUROC difference shows a trend of first decreasing and then increasing, reaching its lowest point at $\phi = 0$. The reason for this phenomenon is that when $\phi = 0$, there is no graph topology information, so different shadow datasets sampling of hard and weak setting do not significantly affect the attack AUROC, as the node features are sampled from the same Gaussian distribution for each class. But as $|\phi|$ increases, more graph topology information is involved in the training process, leading to a larger difference in shadow datasets distributions between hard and weak setting, which results in a significant difference in MIA attack intensity. This can be reflected by the larger disparity in attack AUROC. This phenomenon indicates that graph topology information will increase the intensity of MIA, making it more challenging to protect the label membership of graph data, and distinguish the graph transductive MIA from its graphless counterpart.

\vspace{-0.1in}
\section{Conclusions and Limitations}
\vspace{-0.05in}
\label{Conclusion}
We proposed a novel two-stage defense method (GTD) against MIA tailored for GNNs, and deployed it in an transductive setting for the first time. We compared the performance of GTD with LBP and DMP and demonstrated that GTD achieves the new state-of-the-art. We conducted ablation studies and validated the origin of GTD's defense capability. We also analyzed how graph topology impacts GTD performance. As GTD exhibits superior performance and is easy to be integrated into various GNNs training, we believe it can be highly practical and widely used in this field. 

\textbf{Limitations and Future Work.} Current version of GTD still has some limitations: (1) It could possibly lead to lower model utility because of the labels used in second stage is psudolabel of test nodes, instead of groudtruth labels; (2) The flattening parameter $\beta$ is not end-to-end learnable, and the uniform flattening may not be the optimal way to counter MIA. To address these limitations, we plan to use only the test nodes with high confidence predictions and change the formula of soft labels to make $\beta$ learnable in the future.

\clearpage
\bibliographystyle{plainnat}

\clearpage
\appendix

\section{Extended Related Works}
\label{app:related}
\textbf{Membership Inference Attacks.} MIA on ML models aim to infer whether a data record was used to train a target ML model or not.  This concept is firstly proposed by~\citet{homer2008resolving} and later on extended to various directions, ranging from white-box setting where the whole
target model is released~\citep{Nasr_2019,rezaei2021difficulty,melis2019exploiting,leino2020stolen}, to black-box setting where only (partial of) output predictions are accessible to the adversary~\citep{shokri2017membership,salem2018mlleaks,song2020systematic,li2021membership,choquette2021label,carlini2022membership}. As a general guideline for MIA, the attacker first need to determine the most informative features that distinguish the sample membership. This feature can be posterior predictions~\citep{shokri2017membership,salem2018mlleaks,jia2019memguard}, loss values~\citep{yeom2018privacy,sablayrolles2019whitebox}, or gradient norms~\citep{Nasr_2019,rezaei2021difficulty}. Upon identifying the informative features, the attacker can choose to learn either a binary classifier~\citep{shokri2017membership} or metric-based decisions~\citep{yeom2018privacy,salem2018mlleaks} from shadow model trained on shadow dataset to extract patterns in these features among the training samples for identifying
membership. The shadow dataset can be either generated from target model inferences, or a noisy version of the original dataset depending on the assumptions of the attacker.

\textbf{Defense Against Membership Inference Attacks.} As MIA exploit the behavioral differences of the target model on trainset and testset, most defense mechanisms work towards suppressing the common patterns that an optimal attack relies on. Popular defense methods include confidence score masking, regularization, knowledge distillation, and differential privacy. Confidence score masking aims to hide the true prediction vector returned by the target model and thus mitigates the effectiveness of MIAs, including only providing top-$k$ logits per inference~\citep{shokri2017membership}, or add noise to the prediction vector in an adversarial manner~\citep{jia2019memguard}. Regularization aims to reduce the overfitting degree of target models to mitigate MIAs. Existing regularization methods including $L_2$-norm regularization~\citep{choquette2021label,hayes2017logan}, dropout~\citep{leino2020stolen,salem2018mlleaks},
data argumentation~\citep{kaya2021does,yu2021does}, model compression~\citep{wang2020against}, and label smoothing~\citep{chen2022relaxloss}. Knowledge distillation aims to transfer the knowledge from a unprotected model to a protected model~\citep{shejwalkar2020membership}, and differential privacy~\citep{saeidian2021quantifying,chien2024differentially} naturally protects the membership information with theoretical guarantees at the cost of lower model utility.


\section{Standard MIA Process}
\label{app:mia_process}
\begin{figure}[h]
  \centering
  \includegraphics[width=\linewidth]{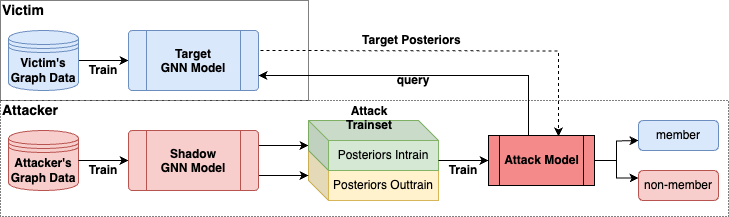}
  \caption{The Process of Membership Inference Attack in graph transductive setting. The shadow GNN model generates attack trainset for training the attack model. Attackers then query the trained attack model with posteriors obtained from the target model (target predictions) to infer membership.
  }
  \label{fig:mia}
\end{figure}
For attackers, their standard MIA process has three phases: shadow GNN model training, attack GNN model training, and membership inference. (1) shadow GNN model training: shadow GNN model $S$ is a model trained by attackers to replicate the behavior of the target GNN model $M$, providing training data for the attack model $A$. To train $S$, we assume that the shadow dataset $\mathcal{G}_{s}$ comes from the same or similar underlying distribution as $\mathcal{G}_{t}$. Then the attackers train $S$ by using $(X_{s}, A_{s}, Y^{\text{Train}}_{s}, \mathcal{V}^{\text{Train}}_{s})$ (2) attack model training: To train $A$, attackers use the trained $S$ to predict all nodes in $\mathcal{V}^{\text{Train}}_{s}$ and $\mathcal{V}^{\text{Test}}_{s}$ and obtain the corresponding posteriors. For each node, attackers take its posteriors as input of the attack model and assigns a label "1" if the node is from $\mathcal{V}^{\text{Train}}_{s}$ and "0" if the node is from $\mathcal{V}^{\text{Test}}_{s}$ to supervise. (3) membership inference: To implement membership inference attack on a given node $v$, attackers query $M$ with $v$'s feature to obtain its posterior. Then attackers input the posterior into the attack model to obtain the membership information.

\section{Details of DMP loss function}
\label{DMP}
In our experiments, the post-distillation phase of DMP consists of two parts of loss to train the protected model, with the proportion adjusted by a hyperparameter. One loss is the cross-entropy loss, supervised by the true labels of the reference data. The other loss is the KL divergence between the prediction of the protected model and the unprotected model on the reference data. The former is to ensure that the protected model has a high classification accuracy on the testset, while the latter is to guide the protected model by using the knowledge from the unprotected model. In our experiments, we adjust the hyperparameters to balance the testset classification accuracy and defense capability of the protected model.

\section{Complete Experimental Results}
\label{app:complete_exp}
Table~\ref{tab:Benchmark} contains properties and statistics about benchmark datasets we used in our experiments. For target models and shadow models, we used 2-layer GCN, 2-layer GAT, 2-layer SGC, NLGCN, NLGAT, NLMLP, and GPRGNN architecture. The attack model is a 3-layer MLP model. The optimizer we used is Adam. All target and shadow models are trained such that they achieve comparable performance as reported by the authors in the literature. We used one NVIDIA GeForce RTX 3090 for training. The time for finishing one experiment is about 10 minutes to 5 hours depends on the complexity of datasets.
\begin{table*}[h]
  \caption{Benchmark dataset properties and statistics: $|\mathcal{V}|$ and $|\mathcal{E}|$ denote the number of vertices and edges in the corresponding graph dataset.}
  \label{tab:Benchmark}
  \centering
  \begin{tabular}{lllll}
    \toprule
    Dataset & $|\mathcal{V}|$ & $|\mathcal{E}|$ & Features & Classes\\
    \midrule
        \ Cora & 2708 & 5278 & 1433 & 7  \\
        \ CiteSeer & 3327 & 4552 & 3703 & 6  \\
        \ PubMed & 19717 & 44324 & 500 & 5  \\
        \ Computers & 13752 & 245861 & 767 & 10 \\
        \ Photo & 7650 & 119081 & 745 & 8  \\
        \ Ogbn-Arxiv & 169343 & 1166243 & 128 & 40  \\
        \ Texas & 183 & 279 & 1703 & 5 \\
        \ Chameleon & 2277 & 31371 & 2325 & 5 \\
        \ Squirrel & 5201 & 198353 & 2089 & 5 \\
    \bottomrule
    \end{tabular}
\end{table*}

\subsection{Comparison with LBP}
\label{Comparison with LBP}
The parameters we used for LBP is shown in Table~\ref{tab:LBP Parameters}. For each experiment, we repeated 5 times and presented the mean and standard deviation of the results in Table~\ref{tab:LBP} and Table~\ref{tab:LBP Weak}. Table~\ref{tab:LBP1} and Table~\ref{tab:LBP1 Weak} show the relative change rates of GTD compared to the LBP defense method on "CLassify Acc" and "Attack AUROC". Note that Table~\ref{tab:LBP} and Table~\ref{tab:LBP1} is the result of experiments in hard setting, Table~\ref{tab:LBP Weak} and Table~\ref{tab:LBP1 Weak} is the result of experiments in weak setting. It can be seen that the result of experiments in weak setting have larger standard deviation due to the randomness of sampling datasets.

Our analysis corresponding to different datasets is as follows:

For Cora and CiteSeer datasets, our defense method has a slightly smaller adverse impact on the target model compared to LBP defense method. However, it exhibits a more significant advantage in defending against attacks, resulting in the attack model's classification accuracy being lower than random classification, effectively eliminating the risk of membership inference attacks. After applying the defense method to the four models, GCN and GAT exhibit similar performance(GAT is a little bit better than GCN), while SGC performs the worst, as it has the highest probability of being successfully attacked. Although GPRGNN also has a high probability of being successfully attacked, its impact on the target model is minimal. As a result, GPRGNN's overfitting problem is the most serious. This is because GPRGNN is too powerful for simple dataset like Cora and CiteSeer, which means GPRGNN has memorized the training dataset excessively. At the same time, GAT demonstrates the best generalization capability(because GAT has the mildest overfitting problem, as the attention mechanism of GAT enables the target model to learn the commonalities between the trainset and testset-both of two datasets have similar node relationships. Therefore, not only does GAT exhibit the strongest generalization capability on the Cora and CiteSeer datasets, but it also has the optimal ability to resist membership inference attacks on other datasets. 

For PubMed, Computers and Photo datasets, GTD achieves much better classify performance. However, there is a slight improvement in defense capability. This is because the average degree of nodes in these three datasets is relatively large, and similar nodes tend to cluster in greater numbers. The target model can learn classification capabilities through a large number of similar node features, leading to more severe overfitting on the testsets, making the attack model more dangerous. Although LBP defense method can also achieve decent defense capability, it comes at the cost of significant loss in target model classification capability. Among the four GNN models, our defense method shows the most significant improvement in classification capability over LBP on SGC, consistent with the results obtained on the Cora and CiteSeer datasets. 

For Ogbn-Arxiv  dataset, which is a large set, GTD, relative to the LBP defense method, achieves comparable defense capabilities without excessively sacrificing the model's classification ability. This may be because large-scale datasets can provide sufficient generalization ability for the target model, making it difficult for attackers to perform membership inference.

For Texas dataset, GTD shows significant improvements in both classification and defense capabilities. This is because the Texas dataset has a smaller number of nodes, leading to insufficient training data for the target model and severe overfitting. However, GTD converts the testset into training data for the target model, greatly enhancing the model's generalization ability and thus strengthening its defense capabilities. In contrast, the LBP defense method excessively sacrifices the model's classification ability, making it difficult to be utilized effectively.

For Chameleon and Squirrelv dataset, it can be seen that even with very low model classification accuracy, the attack model can still achieve membership inference with a probability exceeding random selection. GTD demonstrates significant improvements in both model classification and defense capabilities on two datasets. We note that NLMLP's defense capability has been greatly enhanced, not only due to the improvement in its generalization ability but also because we did not excessively sacrifice its classification capability.

\begin{table}[htbp]
  \caption{Parameters for LBP}
  \label{tab:LBP Parameters}
  \centering
  \begin{tabular}{lll}
    \toprule
    Dataset & $N$ & $b$ \\
    \midrule
        \ Cora & 2 & 1 \\
        \ CiteSeer & 2 & 1 \\
        \ PubMed & 2 & 1 \\
        \ Computers & 2 & 0.2 \\
        \ Photo & 2 & 0.2 \\
        \ Ogbn-Arxiv & 2 & 10  \\
        \ Texas & 2 & 0.2 \\
        \ Chameleon & 2 & 0.2 \\
        \ Squirrel & 2 & 0.2 \\
    \bottomrule
    \end{tabular}
\end{table}

\begin{table}
   \caption{Comparing GTD with LBP. Compared to LBP, GTD achieves a decrease in attack AUROC by $9.42\%$ and an increase in utility performance by $18.08\%$ on average.} 
   \label{tab:LBP}
   \centering
   \resizebox{\textwidth}{!}{
   \begin{tabular}{llllll}
    \toprule
    Dataset & Models & Classify Acc(LBP) &Classify Acc(GTD) &Attack AUROC(LBP) &Attack AUROC(GTD)\\
    \midrule
    \multirow{4}{*}{Cora}
    & GCN& 0.8321 $\pm$ 0.0052& 0.8684 $\pm$ 0.0031 & 0.5115 $\pm$ 0.0043& 0.4659 $\pm$ 0.0042\\
    & GAT & 0.8341 $\pm$ 0.0043& 0.8657 $\pm$ 0.0018 & 0.5196 $\pm$ 0.0057& 0.4522 $\pm$ 0.0056\\
    & SGC & 0.8321 $\pm$ 0.0065& 0.8731 $\pm$ 0.0054 & 0.5216 $\pm$ 0.0072& 0.4977 $\pm$ 0.0068\\
    & GPRGNN& 0.8601 $\pm$ 0.0050& 0.8806 $\pm$ 0.0030 & 0.5287 $\pm$ 0.0060& 0.4894 $\pm$ 0.0058\\
    \midrule
    \multirow{4}{*}{CiteSeer} 
    & GCN& 0.6980 $\pm$ 0.0047 & 0.7233 $\pm$ 0.0035 & 0.5138 $\pm$ 0.0060 & 0.4035 $\pm$ 0.0061\\
    & GAT& 0.6952 $\pm$ 0.0051 & 0.7210 $\pm$ 0.0042 & 0.5190 $\pm$ 0.0047 & 0.3789 $\pm$ 0.0049\\
    & SGC& 0.7013 $\pm$ 0.0074 & 0.7362 $\pm$ 0.0065 & 0.5268 $\pm$ 0.0081 & 0.4499 $\pm$ 0.0069\\
    & GPRGNN& 0.7210 $\pm$ 0.0058& 0.7358 $\pm$ 0.0047 & 0.5288 $\pm$ 0.0065& 0.4619 $\pm$ 0.0062\\
    \midrule
    \multirow{4}{*}{PubMed} 
    & GCN& 0.6886 $\pm$ 0.0041 & 0.8381 $\pm$ 0.0023 & 0.4998 $\pm$ 0.0050 & 0.4990 $\pm$ 0.0048\\
    & GAT & 0.7631 $\pm$ 0.0037 & 0.8400 $\pm$ 0.0028 & 0.5021 $\pm$ 0.0084 & 0.4911 $\pm$ 0.0061\\
    & SGC & 0.6564 $\pm$ 0.0035 & 0.8080 $\pm$ 0.0020 & 0.5007 $\pm$ 0.0065 & 0.5005 $\pm$ 0.0057\\
    & GPRGNN& 0.7843 $\pm$ 0.0029 & 0.8553 $\pm$ 0.0014 & 0.5003 $\pm$ 0.0038 & 0.4967 $\pm$ 0.0034\\
    \midrule
    \multirow{4}{*}{Computers} 
    & GCN& 0.6900 $\pm$ 0.0023 & 0.8818 $\pm$ 0.0018 & 0.5128 $\pm$ 0.0039 & 0.5068 $\pm$ 0.0035\\
    & GAT & 0.7414 $\pm$ 0.0026 & 0.9086 $\pm$ 0.0025 & 0.5165 $\pm$ 0.0053 & 0.5039 $\pm$ 0.0051\\ 
    & SGC & 0.6383 $\pm$ 0.0038 & 0.8310 $\pm$ 0.0023 & 0.5112 $\pm$ 0.0041 & 0.5074 $\pm$ 0.0043\\
    & GPRGNN& 0.7203 $\pm$ 0.0023 & 0.8942 $\pm$ 0.0012 & 0.5154 $\pm$ 0.0033 & 0.5050 $\pm$ 0.0030\\ 
    \midrule
    \multirow{4}{*}{Photo} 
    & GCN & 0.7737 $\pm$ 0.0034& 0.9299 $\pm$ 0.0023 & 0.5179 $\pm$ 0.0046& 0.5110 $\pm$ 0.0041\\
    & GAT& 0.8051 $\pm$ 0.0040& 0.9453 $\pm$ 0.0029 & 0.5123 $\pm$ 0.0045& 0.5066 $\pm$ 0.0039\\
    & SGC& 0.7361 $\pm$ 0.0039& 0.9001 $\pm$ 0.0028 & 0.5159 $\pm$ 0.0039& 0.5106 $\pm$ 0.0035\\
    & GPRGNN& 0.8187 $\pm$ 0.0027& 0.9430 $\pm$ 0.0015 & 0.5153 $\pm$ 0.0025& 0.5088 $\pm$ 0.0028\\
    \midrule
    \multirow{4}{*}{Ogbn-Arxiv} 
    &GCN& 0.5097 $\pm$ 0.0025 & 0.7200 $\pm$ 0.0019 & 0.5005 $\pm$ 0.0037 & 0.4995 $\pm$ 0.0032\\
    &GAT& 0.5134 $\pm$ 0.0030 & 0.7223 $\pm$ 0.0024 & 0.5001 $\pm$ 0.0045 & 0.4978 $\pm$ 0.0038\\
    &SGC& 0.5021 $\pm$ 0.0032 & 0.7272 $\pm$ 0.0027 & 0.4998 $\pm$ 0.0040 & 0.4923 $\pm$ 0.0034\\
    &GPRGNN& 0.5221 $\pm$ 0.0021 & 0.7315 $\pm$ 0.0016 & 0.5017 $\pm$ 0.0031 & 0.5002 $\pm$ 0.0026\\
    \midrule
    \multirow{4}{*}{Texas} 
    & NLGCN& 0.5113 $\pm$ 0.0033& 0.6152 $\pm$ 0.0031 & 0.5954 $\pm$ 0.0061& 0.4812 $\pm$ 0.0054\\
    & NLGAT& 0.5327 $\pm$ 0.0037& 0.5882 $\pm$ 0.0026 & 0.5710 $\pm$ 0.0045& 0.3885 $\pm$ 0.0043\\
    & NLMLP& 0.5713 $\pm$ 0.0062& 0.6686 $\pm$ 0.0043 & 0.5585 $\pm$ 0.0035 & 0.5532 $\pm$ 0.0038\\
    & GPRGNN& 0.6166 $\pm$ 0.0041& 0.7224 $\pm$ 0.0029 & 0.6201 $\pm$ 0.0038& 0.4559 $\pm$ 0.0032\\
    \midrule
    \multirow{4}{*}{Chameleon} 
    & NLGCN& 0.5987 $\pm$ 0.0068&0.6657 $\pm$ 0.0062&0.5033 $\pm$ 0.0062&0.4854 $\pm$ 0.0065\\
    & NLGAT& 0.5926 $\pm$ 0.0072&0.6585 $\pm$ 0.0070&0.5046 $\pm$ 0.0065&0.4602 $\pm$ 0.0063\\
    & NLMLP& 0.4281 $\pm$ 0.0078&0.4824 $\pm$ 0.0074&0.5523 $\pm$ 0.0057&0.3048 $\pm$ 0.0051\\
    & GPRGNN& 0.5230 $\pm$ 0.0054&0.6550 $\pm$ 0.0058&0.5107 $\pm$ 0.0060&0.4936 $\pm$ 0.0049\\
    \midrule
    \multirow{4}{*}{Squirrel} 
    &NLGCN&0.4056 $\pm$ 0.0078&0.4910 $\pm$ 0.0073&0.5039 $\pm$ 0.0059&0.4913 $\pm$ 0.0060\\
    &NLGAT&0.4503 $\pm$ 0.0082&0.5446 $\pm$ 0.0077&0.5160 $\pm$ 0.0053&0.4420 $\pm$ 0.0049 \\
    &NLMLP& 0.2964 $\pm$ 0.0085&0.3137 $\pm$ 0.0068&0.5659 $\pm$ 0.0058&0.3046 $\pm$ 0.0054\\
    &GPRGNN&0.3449 $\pm$ 0.0066&0.4013 $\pm$ 0.0060&0.5125 $\pm$ 0.0055&0.4822 $\pm$ 0.0050\\
    \bottomrule
    \end{tabular}}
\end{table}

\begin{table*}
   \caption{Comparing GTD with LBP. Compared to LBP, GTD achieves a decrease in attack AUROC by $9.42\%$ and an increase in utility performance by $18.08\%$ on average.}
   \label{tab:LBP1}
   \resizebox{\textwidth}{!}{
   \begin{tabular}{llllll}
    \toprule
    Dataset & Models & Classify Acc (LBP) &Classify Acc (GTD) &Attack AUROC (LBP) &Attack AUROC (GTD)\\
    \midrule
    \multirow{4}{*}{Cora}& GCN& 0.8321 & 0.8684 (+4.63\%)& 0.5115 & 0.4659(-8.91\%)\\
    & GAT & 0.8341 & 0.8657(+3.79\%)& 0.5196 & 0.4522(-12.97\%)\\
    & SGC & 0.8321 & 0.8731(+4.93\%)& 0.5216 & 0.4977(-4.58\%)\\
    & GPRGNN& 0.8601 & 0.8806(+2.38\%)& 0.5287 & 0.4894(7.43\%)\\
    \midrule
    \multirow{4}{*}{CiteSeer} & GCN& 0.6980 & 0.7233(+3.62\%)& 0.5138 & 0.4035(-21.47\%)\\
    & GAT& 0.6952 & 0.7210(+3.71\%)& 0.5190 & 0.3789(-26.99\%)\\
    & SGC& 0.7013 & 0.7362(+4.98\%)& 0.5268 & 0.4499(-14.60\%)\\
    & GPRGNN& 0.7210 & 0.7358(+2.05\%)& 0.5288 & 0.4619(-12.65\%)\\
    \midrule
    \multirow{4}{*}{PubMed} & GCN& 0.6886 & 0.8381(+21.71\%)& 0.4998 & 0.4990(-0.16\%)\\
    & GAT & 0.7631 & 0.8400(+10.08\%)& 0.5021 & 0.4911(-2.19\%)\\
    & SGC & 0.6564 & 0.8080(+23.10\%)& 0.5007 & 0.5005(-0.04\%)\\
    & GPRGNN& 0.7843 & 0.8553(+9.05\%)& 0.5003 & 0.4967(-0.72\%)\\
    \midrule
    \multirow{4}{*}{Computers} & GCN& 0.6900 & 0.8818(+27.80\%)& 0.5128 & 0.5068(-1.17\%)\\
    & GAT & 0.7414 & 0.9086(+22.55\%)& 0.5165 & 0.5039(-2.44\%)\\ 
    & SGC & 0.6383 & 0.8310(+30.19\%)& 0.5112 & 0.5074(-0.74\%)\\
    & GPRGNN& 0.7203 & 0.8942(+24.14\%)& 0.5154 & 0.5050(-2.02\%)\\ 
    \midrule
    \multirow{4}{*}{Photo} & GCN & 0.7737 & 0.9299(+19.93\%)& 0.5179 & 0.5110(-1.33\%)\\
    & GAT& 0.8051 & 0.9453(+17.41\%)& 0.5123 & 0.5066(-1.11\%)\\
    & SGC& 0.7361 & 0.9001(+22.28\%)& 0.5159 & 0.5106(-1.03\%)\\
    & GPRGNN& 0.8187 & 0.9430(+15.18\%)& 0.5153 & 0.5088(-1.26\%)\\
    \midrule
    \multirow{4}{*}{Ogbn-Arxiv} 
    &GCN& 0.5097 & 0.7200(+41.26\%)& 0.5005 & 0.4995(-0.20\%)\\
    &GAT& 0.5134 & 0.7223(+40.69\%)& 0.5001 & 0.4978(-0.46\%)\\
    &SGC& 0.5021 & 0.7272(+44.83\%)& 0.4998 & 0.4923(-1.50\%)\\
    &GPRGNN& 0.5221 & 0.7315(+40.11\%)& 0.5017 & 0.5002(-0.30\%)\\
    \midrule
    \multirow{4}{*}{Texas} 
    & NLGCN& 0.5113 & 0.6152(+20.32\%)& 0.5954 & 0.4812(-23.72\%)\\
    & NLGAT& 0.5327 & 0.5882(+10.46\%)& 0.5710 & 0.3885(-31.96\%)\\
    & NLMLP& 0.5713 & 0.6686(+17.02\%)& 0.5585 & 0.5532(-0.95\%)\\
    & GPRGNN& 0.6166 & 0.7224(+17.15\%)& 0.6201 & 0.4559(-26.48\%)\\
    \midrule
    \multirow{4}{*}{Chameleon} 
    & NLGCN&0.5987&0.6657(+12.35\%)&0.5033&0.4854(-3.56\%)\\
    & NLGAT& 0.5926&0.6585(+11.12\%)&0.5046&0.4602(-8.80\%)\\
    & NLMLP& 0.4281&0.4824(+12.68\%)&0.5523&0.3048(-44.81\%)\\
    & GPRGNN& 0.5230&0.6550(+25.24\%)&0.5107&0.4936(-3.35\%)\\
    \midrule
    \multirow{4}{*}{Squirrel} &NLGCN&0.4056&0.4910(+21.06\%)&0.5039&0.4913(-2.50\%)\\
    &NLGAT&0.4503&0.5446(+20.87\%)&0.5160&0.4420(-14.34\%)\\
    &NLMLP& 0.2964&0.3137(+5.84\%)&0.5659&0.3046(-46.17\%)\\
    &GPRGNN&0.3449&0.4013(+16.35\%)&0.5125&0.4822(-6.34\%)\\
    \bottomrule
    \end{tabular}}
\end{table*}

\begin{table*}
  \caption{Comparing GTD with LBP in weak setting.}
  \label{tab:LBP Weak}
  \resizebox{\textwidth}{!}{
  \begin{tabular}{llllll}
    \toprule
    Dataset & Models & Classify Acc (LBP) & Classify Acc (GTD) & Attack AUROC (LBP) & Attack AUROC (GTD) \\
    \midrule 
     \multirow{4}{*}{Cora}
     & GCN & 0.8433 $\pm$ 0.0061& 0.8792 $\pm$ 0.0050 & 0.4832 $\pm$ 0.0070& 0.4385 $\pm$ 0.0073 \\
     & GAT & 0.8429 $\pm$ 0.0052& 0.8718 $\pm$ 0.0042 & 0.4928 $\pm$ 0.0083& 0.4339 $\pm$ 0.0076 \\
     & SGC & 0.8411 $\pm$ 0.0070& 0.8793 $\pm$ 0.0062 & 0.5011 $\pm$ 0.0092& 0.4729 $\pm$ 0.0089 \\
     & GPRGNN & 0.8699 $\pm$ 0.0063 & 0.8871 $\pm$ 0.0048 & 0.5120 $\pm$ 0.0078& 0.4732 $\pm$ 0.0074 \\
    \midrule
     \multirow{4}{*}{CiteSeer}
     & GCN & 0.7198 $\pm$ 0.0054& 0.7366 $\pm$ 0.0056 & 0.5068 $\pm$ 0.0079& 0.4015 $\pm$ 0.0072 \\
     & GAT & 0.7061 $\pm$ 0.0069& 0.7327 $\pm$ 0.0060 & 0.5031 $\pm$ 0.0068& 0.3814 $\pm$ 0.0075 \\
     & SGC & 0.7109 $\pm$ 0.0066& 0.7491 $\pm$ 0.0068 & 0.5223 $\pm$ 0.0095& 0.4433 $\pm$ 0.0093 \\
     & GPRGNN & 0.7323 $\pm$ 0.0068& 0.7486 $\pm$ 0.0054 & 0.5258 $\pm$ 0.0076& 0.4580 $\pm$ 0.0079 \\
    \midrule
    \multirow{4}{*}{PubMed}
     & GCN & 0.6843 $\pm$ 0.0060& 0.8344 $\pm$ 0.0061 & 0.4989 $\pm$ 0.0070& 0.4976 $\pm$ 0.0075 \\
     & GAT & 0.7601 $\pm$ 0.0052& 0.8378 $\pm$ 0.0048 & 0.5016 $\pm$ 0.0068& 0.4901 $\pm$ 0.0066 \\
     & SGC & 0.6532 $\pm$ 0.0058& 0.8046 $\pm$ 0.0049 & 0.5012 $\pm$ 0.0082& 0.5000 $\pm$ 0.0084 \\
     & GPRGNN & 0.7793 $\pm$ 0.0047& 0.8391 $\pm$ 0.0034 & 0.5008 $\pm$ 0.0059& 0.4977 $\pm$ 0.0063 \\
    \midrule
    \multirow{4}{*}{Chameleon}
     & NLGCN & 0.6031 $\pm$ 0.0083& 0.6723 $\pm$ 0.0080 & 0.4961 $\pm$ 0.0089& 0.4729 $\pm$ 0.0086 \\
     & NLGAT & 0.6032 $\pm$ 0.0092& 0.6644 $\pm$ 0.0085 & 0.4993 $\pm$ 0.0092& 0.4552 $\pm$ 0.0079 \\
     & NLMLP & 0.4400 $\pm$ 0.0089& 0.4911 $\pm$ 0.0081 & 0.5520 $\pm$ 0.0087& 0.3129 $\pm$ 0.0071 \\
     & GPRGNN & 0.5340 $\pm$ 0.0079& 0.6556 $\pm$ 0.0076 & 0.4900 $\pm$ 0.0075& 0.4683 $\pm$ 0.0069 \\
    \bottomrule
  \end{tabular}}
\end{table*}

\begin{table*}
  \caption{Comparing GTD with LBP in weak setting.}
  \label{tab:LBP1 Weak}
  \resizebox{\textwidth}{!}{
  \begin{tabular}{llllll}
    \toprule
    Dataset & Models & Classify Acc (LBP) & Classify Acc (GTD) & Attack AUROC (LBP) & Attack AUROC (GTD) \\
    \midrule 
     \multirow{4}{*}{Cora}
     & GCN & 0.8433 & 0.8792 (+4.26\%) & 0.4832 & 0.4385 (-9.25\%) \\
     & GAT & 0.8429 & 0.8718 (+3.43\%) & 0.4928 & 0.4339 (-11.95\%) \\
     & SGC & 0.8411 & 0.8793 (+4.54\%) & 0.5011 & 0.4729 (-5.63\%) \\
     & GPRGNN & 0.8699 & 0.8871 (+1.98\%) & 0.5120 & 0.4732 (-7.58\%) \\
    \midrule
     \multirow{4}{*}{CiteSeer}
     & GCN & 0.7198 & 0.7366 (+2.33\%) & 0.5068 & 0.4015 (-20.78\%) \\
     & GAT & 0.7061 & 0.7327 (+3.77\%) & 0.5031 & 0.3814 (-24.19\%) \\
     & SGC & 0.7109 & 0.7491 (+5.37\%) & 0.5223 & 0.4433 (-15.13\%) \\
     & GPRGNN & 0.7323 & 0.7486 (+2.23\%) & 0.5258 & 0.4580 (-12.89\%) \\
    \midrule
    \multirow{4}{*}{PubMed}
     & GCN & 0.6843 & 0.8344 (+21.93\%) & 0.4989 & 0.4976 (-0.26\%) \\
     & GAT & 0.7601 & 0.8378 (+10.22\%) & 0.5016 & 0.4901 (-2.29\%) \\
     & SGC & 0.6532 & 0.8046 (+23.18\%) & 0.5012 & 0.5000 (-0.24\%) \\
     & GPRGNN & 0.7793 & 0.8391 (+7.67\%) & 0.5008 & 0.4977 (-0.62\%) \\
    \midrule
    \multirow{4}{*}{Chameleon}
     & NLGCN & 0.6031 & 0.6723 (+11.47\%) & 0.4961 & 0.4729 (-4.68\%) \\
     & NLGAT & 0.6032 & 0.6644 (+10.15\%) & 0.4993 & 0.4552 (-8.83\%) \\
     & NLMLP & 0.4400 & 0.4911 (+11.61\%) & 0.5520 & 0.3129 (-43.32\%) \\
     & GPRGNN & 0.5340 & 0.6556 (+22.77\%) & 0.4900 & 0.4683 (-4.43\%) \\
    \bottomrule
  \end{tabular}}
\end{table*}

\subsection{Comparison with DMP}
\label{Comparison with DMP}
For each experiment, we repeated 5 times and presented the mean and standard deviation of the results in Table~\ref{tab:DMP} and Table~\ref{tab:DMP Weak}. Table~\ref{tab:DMP1} and Table~\ref{tab:DMP1 Weak} show the relative change rates of GTD compared to the LBP defense method on "CLassify Acc" and "Attack AUROC". Note that Table~\ref{tab:DMP} and Table~\ref{tab:DMP1} is the result of experiments in hard setting, Table~\ref{tab:DMP Weak} and Table~\ref{tab:DMP1 Weak} is the result of experiments in weak setting. It can be seen that the result of experiments in weak setting have larger standard deviation due to the randomness of sampling datasets.

Our analysis corresponding to different datasets is as follows:

For Cora, CiteSeer and Texas datasets, GTD outperforms the DMP significantly in both testset classification accuracy and defense capability. These three datasets have a small number of nodes, which means that if the DMP method is applied, the model's classification ability will be greatly impaired due to the need to provide reference data for protected target model.  This is the challenging issue faced by the DMP when applied to GNN models. Additionally, the defense effectiveness of the DMP is also inferior to the GTD because the DMP relies on the unprotected model to guide the training of the protected model to improve generalization, which is not as direct as using the testset for training in the GTD.

For PubMed, Computers, Photo and Ogbn-Arxiv datasets, compared to the DMP, GTD has a slight lead in both classification accuracy and defense performance. As the number of nodes in the dataset increases, the knowledge distillation of the DMP method becomes more pronounced in guiding the protected target model, and its defense capability is comparable to the GTD. However, the DMP method still results in a reduction in the amount of training data, which still has a significant negative impact on the model's classification ability.

In the experiments, we also observed that controlling the hyperparameters that determine the proportions of the two different losses in the post distillation phase of DMP is crucial. It requires achieving a tradeoff between classification accuracy and defense capability. Adjusting these hyperparameters will increase the implementation cost of the DMP method.

\begin{table*}[htbp]
  \caption{Comparing GTD with DMP. Compared to DMP, GTD achieves a decrease in attack AUROC by $4.98\%$ and an increase in utility performance by $5.82\%$ on average.}
  \centering
  \label{tab:DMP}
  \resizebox{\textwidth}{!}{
  \begin{tabular}{llllll}
    \toprule
    Dataset & Models & Classify Acc (DMP) & Classify Acc (GTD) & Attack AUROC (DMP) & Attack AUROC (GTD) \\
    \midrule
    \multirow{4}{*}{Cora}
     & GCN & 0.7646 $\pm$ 0.0058 & 0.8842 $\pm$ 0.0058 & 0.5136 $\pm$ 0.0051& 0.5043 $\pm$ 0.0060 \\
     & GAT & 0.7452 $\pm$ 0.0050 & 0.8858 $\pm$ 0.0047 & 0.5104 $\pm$ 0.0053& 0.5005 $\pm$ 0.0049 \\
     & SGC & 0.7521 $\pm$ 0.0071& 0.8902 $\pm$ 0.0041 & 0.5096 $\pm$ 0.0046& 0.4994 $\pm$ 0.0051 \\
     & GPRGNN & 0.6323 $\pm$ 0.0048& 0.8926 $\pm$ 0.0024 & 0.5162 $\pm$ 0.0040& 0.5044 $\pm$ 0.0043 \\
    \midrule
    \multirow{4}{*}{CiteSeer}
     & GCN & 0.7434 $\pm$ 0.0058& 0.7628 $\pm$ 0.0062 & 0.5202 $\pm$ 0.0053& 0.4811 $\pm$ 0.0045 \\
     & GAT & 0.7156 $\pm$ 0.0061& 0.7564 $\pm$ 0.0059 & 0.5176 $\pm$ 0.0059& 0.4766 $\pm$ 0.0045 \\
     & SGC & 0.7403 $\pm$ 0.0068& 0.7692 $\pm$ 0.0049 & 0.5178 $\pm$ 0.0050& 0.4814 $\pm$ 0.0041 \\
     & GPRGNN & 0.7426 $\pm$ 0.0039& 0.7726 $\pm$ 0.0028 & 0.5204 $\pm$ 0.0045& 0.4962 $\pm$ 0.0038 \\
    \midrule
    \multirow{4}{*}{PubMed}
     & GCN & 0.8235 $\pm$ 0.0037& 0.8387 $\pm$ 0.0034 & 0.5026 $\pm$ 0.0039& 0.4978 $\pm$ 0.0042 \\
     & GAT & 0.8027 $\pm$ 0.0047& 0.8434 $\pm$ 0.0024 & 0.5013 $\pm$ 0.0036& 0.5005 $\pm$ 0.0043 \\
     & SGC & 0.8013 $\pm$ 0.0041& 0.8096 $\pm$ 0.0045 & 0.5024 $\pm$ 0.0042& 0.5003 $\pm$ 0.0038 \\
     & GPRGNN & 0.8104 $\pm$ 0.0031& 0.8423 $\pm$ 0.0036 & 0.5020 $\pm$ 0.0027& 0.4994 $\pm$ 0.0023 \\
    \midrule
    \multirow{4}{*}{Computers}
     & GCN & 0.8756 $\pm$ 0.0038& 0.8808 $\pm$ 0.0040 & 0.5055 $\pm$ 0.0042& 0.5004 $\pm$ 0.0045 \\
     & GAT & 0.9071 $\pm$ 0.0044& 0.9127 $\pm$ 0.0038 & 0.5097 $\pm$ 0.0045& 0.4933 $\pm$ 0.0051 \\
     & SGC & 0.8323 $\pm$ 0.0042& 0.8434 $\pm$ 0.0040 & 0.5086 $\pm$ 0.0045& 0.5023 $\pm$ 0.0047 \\
     & GPRGNN & 0.8683 $\pm$ 0.0029& 0.8898 $\pm$ 0.0021 & 0.5045 $\pm$ 0.0020& 0.5036 $\pm$ 0.0032 \\
    \midrule
    \multirow{4}{*}{Photo}
     & GCN & 0.9228 $\pm$ 0.0046& 0.9304 $\pm$ 0.0042 & 0.5061 $\pm$ 0.0052& 0.5004 $\pm$ 0.0048\\
     & GAT & 0.9415 $\pm$ 0.0052& 0.9493 $\pm$ 0.0038 & 0.5072 $\pm$ 0.0058& 0.4966 $\pm$ 0.0055 \\
     & SGC & 0.8933 $\pm$ 0.0042& 0.8988 $\pm$ 0.0037 & 0.5105 $\pm$ 0.0060& 0.5018 $\pm$ 0.0059 \\
     & GPRGNN & 0.9215 $\pm$ 0.0032& 0.9315 $\pm$ 0.0026 & 0.5043 $\pm$ 0.0044& 0.4976 $\pm$ 0.0042\\
    \midrule
    \multirow{4}{*}{Ogbn-Arxiv}
    & GCN & 0.6876 $\pm$ 0.0039& 0.6921 $\pm$ 0.0020 & 0.4920 $\pm$ 0.0045& 0.4852 $\pm$ 0.0049 \\
     & GAT & 0.6869 $\pm$ 0.0037& 0.6907 $\pm$ 0.0018 & 0.4913 $\pm$ 0.0042& 0.4834 $\pm$ 0.0048\\
     & SGC & 0.6798 $\pm$ 0.0023& 0.6884 $\pm$ 0.0023 & 0.4924 $\pm$ 0.0048& 0.4791 $\pm$ 0.0043 \\
     & GPRGNN & 0.6903 $\pm$ 0.0025& 0.6993 $\pm$ 0.0020 & 0.5035 $\pm$ 0.0037& 0.4972 $\pm$ 0.0040 \\
     \midrule
    \multirow{4}{*}{Texas}
     & NLGCN & 0.6846 $\pm$ 0.0069& 0.7027 $\pm$ 0.0044 & 0.4966 $\pm$ 0.0060& 0.4920 $\pm$ 0.0056\\
     & NLGAT & 0.6916 $\pm$ 0.0074& 0.7263 $\pm$ 0.0038 & 0.4923 $\pm$ 0.0063& 0.4276 $\pm$ 0.0058 \\
     & NLMLP & 0.6948 $\pm$ 0.0070& 0.7282 $\pm$ 0.0042 & 0.4926 $\pm$ 0.0068& 0.4253 $\pm$ 0.0063 \\
     & GPRGNN & 0.6115 $\pm$ 0.0054& 0.7445 $\pm$ 0.0031 & 0.5187 $\pm$ 0.0052& 0.5032 $\pm$ 0.0046 \\
    \midrule
    \multirow{4}{*}{Chameleon}
     & NLGCN & 0.6681 $\pm$ 0.0064& 0.6963 $\pm$ 0.0065 & 0.5210 $\pm$ 0.0064& 0.5182 $\pm$ 0.0062 \\
     & NLGAT & 0.6516 $\pm$ 0.0066& 0.7082 $\pm$ 0.0073 & 0.5116 $\pm$ 0.0061& 0.5159 $\pm$ 0.0059 \\
     & NLMLP & 0.4643 $\pm$ 0.0079& 0.4955 $\pm$ 0.0070 & 0.5054 $\pm$ 0.0053& 0.3817 $\pm$ 0.0056 \\
     & GPRGNN & 0.6471 $\pm$ 0.0060& 0.6934 $\pm$ 0.0068 & 0.5172 $\pm$ 0.0067& 0.5163 $\pm$ 0.0045\\
    \midrule
    \multirow{4}{*}{Squirrel}
     & NLGCN & 0.5011 $\pm$ 0.0074& 0.5102 $\pm$ 0.0076 & 0.5053 $\pm$ 0.0064& 0.5001 $\pm$ 0.0058 \\
     & NLGAT & 0.5502 $\pm$ 0.0087& 0.5723 $\pm$ 0.0071 & 0.5146 $\pm$ 0.0058& 0.4834 $\pm$ 0.0055 \\
     & NLMLP & 0.3240 $\pm$ 0.0090& 0.3393 $\pm$ 0.0073 & 0.5035 $\pm$ 0.0055& 0.2098 $\pm$ 0.0050 \\
     & GPRGNN & 0.4424 $\pm$ 0.0061& 0.4581 $\pm$ 0.0068 & 0.5102 $\pm$ 0.0051& 0.5054 $\pm$ 0.0052 \\
    \bottomrule
  \end{tabular}}
\end{table*}

\begin{table*}[htbp]
  \caption{Comparing GTD with DMP. Compared to DMP, GTD achieves a decrease in attack AUROC by $4.98\%$ and an increase in utility performance by $5.82\%$ on average.}
  \label{tab:DMP1}
  \resizebox{\textwidth}{!}{
  \begin{tabular}{llllll}
    \toprule
    Dataset & Models & Classify Acc (DMP) & Classify Acc (GTD) & Attack AUROC (DMP) & Attack AUROC (GTD) \\
    \midrule
    \multirow{4}{*}{Cora}
     & GCN & 0.7646 & 0.8842 (+15.64\%) & 0.5136 & 0.5043 (-1.81\%) \\
     & GAT & 0.7452 & 0.8858 (+18.87\%) & 0.5104 & 0.5005 (-1.94\%) \\
     & SGC & 0.7521 & 0.8902 (+18.36\%) & 0.5096 & 0.4994 (-2.00\%) \\
     & GPRGNN & 0.6323 & 0.8926 (+41.17\%) & 0.5162 & 0.5044 (-2.29\%) \\
    \midrule
    \multirow{4}{*}{CiteSeer}
     & GCN & 0.7434 & 0.7628 (+2.61\%) & 0.5202 & 0.4811 (-7.52\%) \\
     & GAT & 0.7156 & 0.7564 (+5.70\%) & 0.5176 & 0.4766 (-7.92\%) \\
     & SGC & 0.7403 & 0.7692 (+3.90\%) & 0.5178 & 0.4814 (-7.03\%) \\
     & GPRGNN & 0.7426 & 0.7726 (+4.04\%) & 0.5204 & 0.4962 (-4.65\%) \\
    \midrule
    \multirow{4}{*}{PubMed}
     & GCN & 0.8235 & 0.8387 (+1.85\%) & 0.5026 & 0.4978 (-0.96\%) \\
     & GAT & 0.8027 & 0.8434 (+5.07\%) & 0.5013 & 0.5005 (-0.16\%) \\
     & SGC & 0.8013 & 0.8096 (+1.04\%) & 0.5024 & 0.5003 (-0.42\%) \\
     & GPRGNN & 0.8104 & 0.8423 (+3.94\%) & 0.5020 & 0.4994 (-0.52\%) \\
    \midrule
    \multirow{4}{*}{Computers}
     & GCN & 0.8756 & 0.8808 (+0.59\%) & 0.5055 & 0.5004 (-1.01\%) \\
     & GAT & 0.9071 & 0.9127 (+0.62\%) & 0.5097 & 0.4933 (-3.22\%) \\
     & SGC & 0.8323 & 0.8434 (+1.33\%) & 0.5086 & 0.5023 (-1.24\%) \\
     & GPRGNN & 0.8683 & 0.8898 (+2.48\%) & 0.5045 & 0.5036 (-0.18\%) \\
    \midrule
    \multirow{4}{*}{Photo}
     & GCN & 0.9228 & 0.9304 (+0.82\%) & 0.5061 & 0.5004 (-1.13\%) \\
     & GAT & 0.9415 & 0.9493 (+0.83\%) & 0.5072 & 0.4966 (-2.09\%) \\
     & SGC & 0.8933 & 0.8988 (+0.62\%) & 0.5105 & 0.5018 (-1.70\%) \\
     & GPRGNN & 0.9215 & 0.9315 (+1.09\%) & 0.5043 & 0.4976 (-1.33\%) \\
    \midrule
    \multirow{4}{*}{Ogbn-Arxiv}
    & GCN & 0.6876 & 0.6921 (+0.65\%) & 0.4920 & 0.4852 (-1.38\%) \\
     & GAT & 0.6869 & 0.6907 (+0.55\%) & 0.4913 & 0.4834 (-1.61\%) \\
     & SGC & 0.6798 & 0.6884 (+1.27\%) & 0.4924 & 0.4791 (-2.70\%) \\
     & GPRGNN & 0.6903 & 0.6993 (+1.30\%) & 0.5035 & 0.4972 (-1.25\%) \\
      \midrule
    \multirow{4}{*}{Texas}
     & NLGCN & 0.6846 & 0.7027 (+2.65\%) & 0.4966 & 0.4920 (-0.93\%) \\
     & NLGAT & 0.6916 & 0.7263 (+5.01\%) & 0.4923 & 0.4276 (-13.14\%) \\
     & NLMLP & 0.6948 & 0.7282 (+4.80\%) & 0.4926 & 0.4253 (-13.66\%) \\
     & GPRGNN & 0.6115 & 0.7445 (+21.75\%) & 0.5187 & 0.5032 (-2.99\%) \\
    \midrule
    \multirow{4}{*}{Chameleon}
     & NLGCN & 0.6681  & 0.6963 (+4.19\%) & 0.5210 &+0.5182 (-0.58\%) \\
     & NLGAT & 0.6516 & 0.7082 (+8.76\%) & 0.5116 & 0.5159 (-0.78\%) \\
     & NLMLP & 0.4643 & 0.4955 (+6.68\%) & 0.5054 & 0.3817 (-24.55\%) \\
     & GPRGNN & 0.6471 & 0.6934 (+7.11\%) & 0.5172 & 0.5163 (-0.19\%) \\
    \midrule
    \multirow{4}{*}{Squirrel}
     & NLGCN & 0.5011 & 0.5102 (+1.80\%) & 0.5053 & 0.5001 (-0.99\%) \\
     & NLGAT & 0.5502 & 0.5723 (+4.00\%) & 0.5146 & 0.4834 (-6.03\%) \\
     & NLMLP & 0.3240 & 0.3393 (+4.63\%) & 0.5035 & 0.2098 (-58.45\%) \\
     & GPRGNN & 0.4424 & 0.4581 (+3.62\%) & 0.5102 & 0.5054 (-0.98\%) \\
    \bottomrule
  \end{tabular}}
\end{table*}

\begin{table*}[htbp]
  \caption{Comparing GTD with DMP in weak setting.}
  \label{tab:DMP Weak}
  \centering
  \resizebox{\textwidth}{!}{
  \begin{tabular}{llllll}
    \toprule
    Dataset & Models & Classify Acc (DMP) & Classify Acc (Our) & Attack AUROC (DMP) & Attack AUROC (Our) \\
    \midrule
    \multirow{4}{*}{Cora}
     & GCN & 0.735 $\pm$ 0.0065& 0.8819 $\pm$ 0.0069 & 0.5034 $\pm$ 0.0074& 0.4943 $\pm$ 0.0076 \\
     & GAT & 0.7261 $\pm$ 0.0068& 0.8826 $\pm$ 0.0070 & 0.5102 $\pm$ 0.0082& 0.4982 $\pm$ 0.0072 \\
     & SGC & 0.7549 $\pm$ 0.0072& 0.8885 $\pm$ 0.0066 & 0.4968 $\pm$ 0.0088& 0.498 $\pm$ 0.0090 \\
     & GPRGNN & 0.5284 $\pm$ 0.0061& 0.8789 $\pm$ 0.0058 & 0.4902 $\pm$ 0.0069& 0.4803 $\pm$ 0.0068 \\
    \midrule
    \multirow{4}{*}{CiteSeer}
     & GCN & 0.7201 $\pm$ 0.0067& 0.7549 $\pm$ 0.0070 & 0.5056 $\pm$ 0.0068& 0.4769 $\pm$ 0.0073 \\
     & GAT & 0.7213 $\pm$ 0.0066& 0.7555 $\pm$ 0.0073 & 0.5059 $\pm$ 0.0074& 0.473 $\pm$ 0.0074 \\
     & SGC & 0.7379 $\pm$ 0.0079& 0.7693 $\pm$ 0.0070 & 0.5085 $\pm$ 0.0078& 0.4768 $\pm$ 0.0088 \\
     & GPRGNN & 0.7135 $\pm$ 0.0051& 0.7673 $\pm$ 0.0048 & 0.5122 $\pm$ 0.0066& 0.4852 $\pm$ 0.0070 \\
    \midrule
    \multirow{4}{*}{PubMed}
     & GCN & 0.8279 $\pm$ 0.0063& 0.8364 $\pm$ 0.0066 & 0.4996 $\pm$ 0.0072& 0.4934 $\pm$ 0.0070 \\
     & GAT & 0.8140 $\pm$ 0.0055& 0.8433 $\pm$ 0.0053 & 0.4999 $\pm$ 0.0068& 0.4962 $\pm$ 0.0065 \\
     & SGC & 0.8015 $\pm$ 0.0052& 0.8090 $\pm$ 0.0058 & 0.5002 $\pm$ 0.0081& 0.4932 $\pm$ 0.0076 \\
     & GPRGNN & 0.8180 $\pm$ 0.0044& 0.8649 $\pm$ 0.0050 & 0.5012 $\pm$ 0.0063& 0.4955 $\pm$ 0.0069 \\
    \midrule
    \multirow{4}{*}{Chameleon}
     & NLGCN & 0.6636 $\pm$ 0.0088& 0.6929 $\pm$ 0.0084 & 0.5112 $\pm$ 0.0083& 0.5016 $\pm$ 0.0078 \\
     & NLGAT & 0.6429 $\pm$ 0.0082& 0.6952 $\pm$ 0.0078 & 0.5119 $\pm$ 0.0087& 0.5088 $\pm$ 0.0075 \\
     & NLMLP & 0.4577 $\pm$ 0.0096& 0.4991 $\pm$ 0.0090 & 0.5025 $\pm$ 0.0078& 0.3666 $\pm$ 0.0068 \\
     & GPRGNN & 0.6440 $\pm$ 0.0078& 0.6859 $\pm$ 0.0083 & 0.5010 $\pm$ 0.0064& 0.4978 $\pm$ 0.0063 \\
    \bottomrule
  \end{tabular}}
\end{table*}

\begin{table*}[htbp]
  \caption{Comparing GTD with DMP in weak setting.}
  \label{tab:DMP1 Weak}
  \resizebox{\textwidth}{!}{
  \begin{tabular}{llllll}
    \toprule
    Dataset & Models & Classify Acc(DMP) & Classify Acc(Our) & Attack AUROC(DMP) & Attack AUROC(Our) \\
    \midrule
    \multirow{4}{*}{Cora}
     & GCN & 0.735 & 0.8819 (+19.99\%) & 0.5034 & 0.4943 (-1.81\%) \\
     & GAT & 0.7261 & 0.8826 (+21.55\%) & 0.5102 & 0.4982 (-2.35\%) \\
     & SGC & 0.7549 & 0.8885 (+17.70\%) & 0.4968 & 0.498 (+0.24\%) \\
     & GPRGNN & 0.5284 & 0.8789 (+66.33\%) & 0.4902 & 0.4803 (-2.02\%) \\
    \midrule
    \multirow{4}{*}{CiteSeer}
     & GCN & 0.7201 & 0.7549 (+4.83\%) & 0.5056 & 0.4769 (-5.68\%) \\
     & GAT & 0.7213 & 0.7555 (+4.74\%) & 0.5059 & 0.473 (-6.50\%) \\
     & SGC & 0.7379 & 0.7693 (+4.26\%) & 0.5085 & 0.4768 (-6.23\%) \\
     & GPRGNN & 0.7135 & 0.7673 (+7.54\%) & 0.5122 & 0.4852 (-5.27\%) \\
    \midrule
    \multirow{4}{*}{PubMed}
     & GCN & 0.8279 & 0.8364 (+1.03\%) & 0.4996 & 0.4934 (-1.24\%) \\
     & GAT & 0.8140 & 0.8433 (+3.60\%) & 0.4999 & 0.4962 (-0.74\%) \\
     & SGC & 0.8015 & 0.8090 (+0.94\%) & 0.5002 & 0.4932 (-1.40\%) \\
     & GPRGNN & 0.8180 & 0.8649 (+5.73\%) & 0.5012 & 0.4955 (-1.14\%) \\
    \midrule
    \multirow{4}{*}{Chameleon}
     & NLGCN & 0.6636 & 0.6929 (+4.42\%) & 0.5112 & 0.5016 (-1.88\%) \\
     & NLGAT & 0.6429 & 0.6952 (+8.14\%) & 0.5119 & 0.5088 (-0.61\%) \\
     & NLMLP & 0.4577 & 0.4991 (+9.05\%) & 0.5025 & 0.3666 (-27.04\%) \\
     & GPRGNN & 0.6440 & 0.6859 (+6.51\%) & 0.5010 & 0.4978 (-0.64\%) \\
    \bottomrule
  \end{tabular}}
\end{table*}

\subsection{GTD Defense Method Reduce the Generalization Gap}
\label{Extra experiment about "Two-Stage Defense Method Reduce the Generalization Gap"}
\begin{figure*}[htbp]
    \centering
    \includegraphics[width=\linewidth]{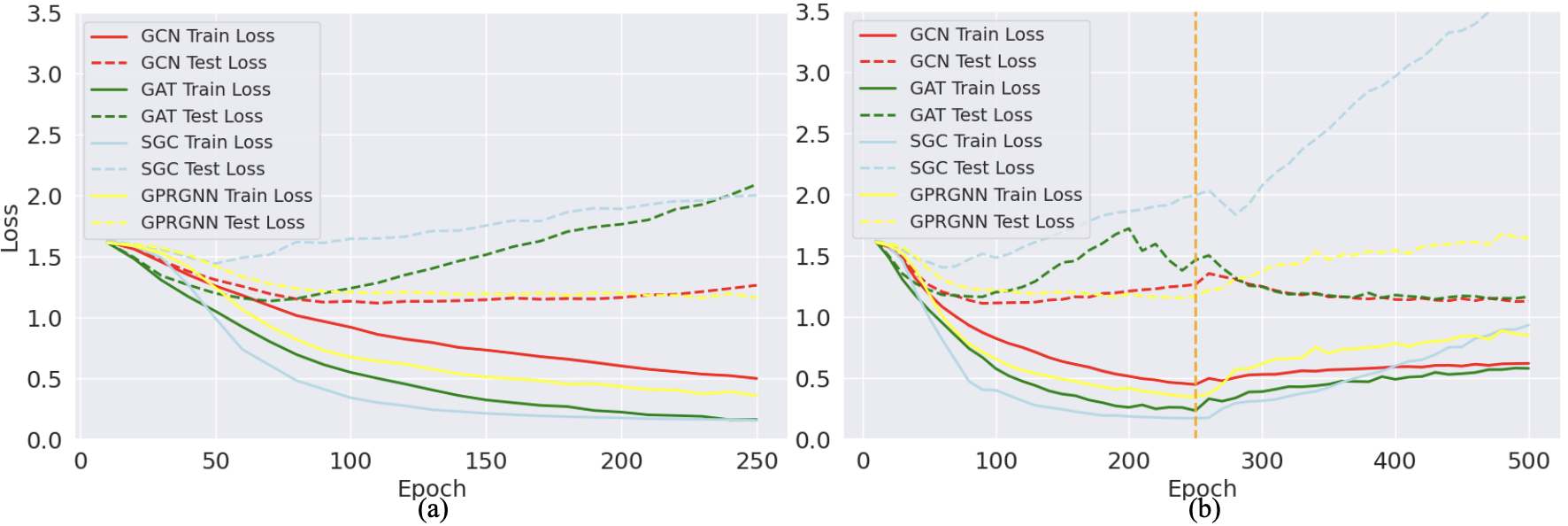}
    \caption{Comparison of average training loss and testing loss on Chameleon for (a) normal training, (b) two-stage training (GTD). In (b), the left half of the orange dashed line indicates the first training stage of our method, while the one on the right indicates the second stage.}
    \label{fig:Chameleon_Train_Test_Loss}
\end{figure*}

Table~\ref{tab:normal vs gtd} contains the average losses of all models from Figure~\ref{fig:Cora_Train_Test_Loss} and Figure~\ref{fig:Chameleon_Train_Test_Loss} at the end of training. Besides, Table~\ref{tab:normal vs gtd} also contains the comparison of normal training and GTD training when facing MIA. The result shows that GTD only slightly decrease the utility of target models, but significantly improve their defense capabilities.

\begin{table*}[htbp]
  \caption{The performance comparison between normal training and GTD training.}
  \centering
  \label{tab:normal vs gtd}
  \resizebox{\textwidth}{!}{
  \begin{tabular}{llllllllll}
    \toprule
    \multirow{2}{*}{Dataset} & \multirow{2}{*}{Models} & \multicolumn{2}{c}{Avg Train Loss} & \multicolumn{2}{c}{Avg Test Loss} & \multicolumn{2}{c}{Classify Acc} & \multicolumn{2}{c}{Attack AUROC}\\ \cline{3-10}
    & &Normal& GTD&Normal& GTD&Normal& GTD&Normal& GTD\\
    \midrule
    \multirow{4}{*}{Cora} 
    & GCN&0.1521&0.4347&0.3901&0.5435&0.8711&0.8684&0.5522&0.4659\\
    & GAT&0.0396&0.4030&0.4585&0.5123&0.8614&0.8657&0.5679&0.4522\\
    & SGC&0.2572&0.4440&0.4049&0.5247&0.8748&0.8731&0.5301&0.4977\\
    & GPRGNN&0.1389&0.3166&0.3481&0.4416&0.8825&0.8806&0.5869&0.4894\\
    \midrule
    \multirow{4}{*}{Chameleon} 
    &NLGCN&0.4988&0.6192&1.1278&1.2631&0.6631&0.6657&0.5267&0.4854\\
    &NLGAT&0.1617&0.5782&1.1650&2.0911&0.6603&0.6585&0.5304&0.4602\\
    &NLMLP&0.1555&0.9322&3.6396&2.0010&0.4857&0.4824&0.7681&0.3048\\  &GPRGNN&0.3575&0.8498&1.6416&1.1626&0.6582&0.6550&0.5837&0.4936\\
    \bottomrule
    \end{tabular}}
\end{table*}

\subsection{Complete Results of the Ablation Study}
\label{Complete Results of the Ablation Study}
\begin{table*}[htbp]
  \caption{Ablation Study}
  \label{tab:Ablation1}
  \resizebox{\textwidth}{!}{
  \begin{tabular}{lllll}
    \toprule
    Method & Dataset & GNN Models & Classify Acc &Attack AUROC \\
    \midrule
    \multirow{16}{*}{Normal Training} 
    & \multirow{4}{*}{Cora} 
    & GCN & 0.8742 $\pm$ 0.0021& 0.5712 $\pm$ 0.0035\\
    & & GAT& 0.8731 $\pm$ 0.0020& 0.5734 $\pm$ 0.0042\\
    & & SGC& 0.8784 $\pm$ 0.0039& 0.5465 $\pm$ 0.0044\\
    & & GPRGNN &0.8856 $\pm$ 0.0031& 0.5479 $\pm$ 0.0051\\
    & \multirow{4}{*}{CiteSeer} 
    & GCN& 0.7296 $\pm$ 0.0034& 0.5979 $\pm$ 0.0048 \\
    & & GAT& 0.7274 $\pm$ 0.0032& 0.6156 $\pm$ 0.0037 \\
    & & SGC& 0.7391 $\pm$ 0.0045& 0.5561 $\pm$ 0.0049 \\
    & & GPRGNN &0.7403 $\pm$ 0.0042& 0.5534 $\pm$ 0.0051 \\
    & \multirow{4}{*}{PubMed} 
    & GCN & 0.8423 $\pm$ 0.0027& 0.5206 $\pm$ 0.0032\\
    & & GAT& 0.8421 $\pm$ 0.0025& 0.5106 $\pm$ 0.0043\\
    & & SGC& 0.8129 $\pm$ 0.0024& 0.5135 $\pm$ 0.0047\\
    & & GPRGNN &0.8637 $\pm$ 0.0026& 0.5288 $\pm$ 0.0045\\
    & \multirow{4}{*}{Chameleon} 
    & NLGCN& 0.6817 $\pm$ 0.0054& 0.5469 $\pm$ 0.0051\\
    & & NLGAT& 0.6451 $\pm$ 0.0053 & 0.5790 $\pm$ 0.0058\\
    & & NLMLP& 0.5077 $\pm$ 0.0060& 0.7868 $\pm$ 0.0049\\
    & & GPRGNN &0.6841 $\pm$ 0.0057& 0.5939 $\pm$ 0.0042\\
    \midrule
    \multirow{16}{*}{Flattenning (One-Stage)} 
    & \multirow{4}{*}{Cora} 
    & GCN & 0.8674 $\pm$ 0.0028& 0.5617 $\pm$ 0.0045\\
    & & GAT& 0.8653 $\pm$ 0.0021& 0.5667 $\pm$ 0.0051\\
    & & SGC& 0.8715 $\pm$ 0.0048& 0.5314 $\pm$ 0.0056\\
    & & GPRGNN &0.8792 $\pm$ 0.0026& 0.5388 $\pm$ 0.0049\\
    & \multirow{4}{*}{CiteSeer} 
    & GCN& 0.7234 $\pm$ 0.0036& 0.5886 $\pm$ 0.0063 \\
    & & GAT& 0.7206 $\pm$ 0.0039& 0.6095 $\pm$ 0.0041 \\
    & & SGC& 0.7346 $\pm$ 0.0061& 0.5477 $\pm$ 0.0058 \\
    & & GPRGNN &0.7356 $\pm$ 0.0041& 0.5466 $\pm$ 0.0050 \\
    & \multirow{4}{*}{PubMed} 
    & GCN & 0.8345 $\pm$ 0.0025& 0.5078 $\pm$ 0.0041\\
    & & GAT& 0.8328 $\pm$ 0.0023& 0.5013 $\pm$ 0.0054\\
    & & SGC& 0.8040 $\pm$ 0.0023& 0.5063 $\pm$ 0.0053\\
    & & GPRGNN &0.8541 $\pm$ 0.0016& 0.5024 $\pm$ 0.0036\\
    & \multirow{4}{*}{Chameleon} 
    & NLGCN& 0.6720 $\pm$ 0.0068& 0.5327 $\pm$ 0.0055\\
    & & NLGAT& 0.6364 $\pm$ 0.0066 & 0.5685 $\pm$ 0.0064\\
    & & NLMLP& 0.4954 $\pm$ 0.0075& 0.7787 $\pm$ 0.0056\\
    & & GPRGNN &0.6757 $\pm$ 0.0053& 0.5833 $\pm$ 0.0039\\
    \midrule    
    \multirow{16}{*}{Flattenning \& Gradient Asent (One-Stage)} & \multirow{4}{*}{Cora} 
    & GCN & 0.8636 $\pm$ 0.0038& 0.5586 $\pm$ 0.0053\\
    & & GAT& 0.8605 $\pm$ 0.0035& 0.5667 $\pm$ 0.0060\\
    & & SGC& 0.8603 $\pm$ 0.0061& 0.5583 $\pm$ 0.0062\\
    & & GPRGNN &0.8751 $\pm$ 0.0033& 0.5376 $\pm$ 0.0052\\
    & \multirow{4}{*}{CiteSeer} 
    & GCN& 0.7221 $\pm$ 0.0046& 0.5857 $\pm$ 0.0070\\
    & & GAT& 0.7209 $\pm$ 0.0043& 0.5946 $\pm$ 0.0063\\
    & & SGC& 0.7341 $\pm$ 0.0065& 0.5437 $\pm$ 0.0069\\
    & & GPRGNN &0.7356 $\pm$ 0.0045& 0.5469 $\pm$ 0.0061\\
    & \multirow{4}{*}{PubMed} 
    & GCN & 0.8310 $\pm$ 0.0043& 0.5082 $\pm$ 0.0050\\
    & & GAT& 0.8430 $\pm$ 0.0040& 0.5001 $\pm$ 0.0074\\
    & & SGC& 0.8190 $\pm$ 0.0066& 0.5089 $\pm$ 0.0070\\
    & & GPRGNN &0.8657 $\pm$ 0.0041& 0.5042 $\pm$ 0.0042 \\
    & \multirow{4}{*}{Chameleon} 
    & NLGCN& 0.6707 $\pm$ 0.0068& 0.5302 $\pm$ 0.0074 \\
    & & NLGAT& 0.6402 $\pm$ 0.0074& 0.5561 $\pm$ 0.0077\\
    & & NLMLP& 0.4933 $\pm$ 0.0075& 0.7741 $\pm$ 0.0063\\
    & & GPRGNN &0.6753 $\pm$ 0.0064& 0.5829 $\pm$ 0.0062\\
    \midrule
    \multirow{16}{*}{Two-Stage (without Flattening)} 
    & \multirow{4}{*}{Cora} 
    & GCN & 0.8706 $\pm$ 0.0030& 0.4834 $\pm$ 0.0045\\
    & & GAT& 0.8610 $\pm$ 0.0024& 0.4591 $\pm$ 0.0054\\
    & & SGC& 0.8724 $\pm$ 0.0057& 0.5044 $\pm$ 0.0059\\
    & & GPRGNN &0.8804 $\pm$ 0.0029& 0.4923 $\pm$ 0.0055\\
    & \multirow{4}{*}{CiteSeer} 
    & GCN& 0.7218 $\pm$ 0.0042& 0.4438 $\pm$ 0.0066\\
    & & GAT& 0.7239 $\pm$ 0.0038& 0.4105 $\pm$ 0.0054\\
    & & SGC& 0.7350 $\pm$ 0.0059& 0.4729 $\pm$ 0.0060\\
    & & GPRGNN &0.7352 $\pm$ 0.0040& 0.4888$\pm$ 0.0058\\
    & \multirow{4}{*}{PubMed} 
    & GCN & 0.8328 $\pm$ 0.0035& 0.5039 $\pm$ 0.0043\\
    & & GAT& 0.8489 $\pm$ 0.0031& 0.4963 $\pm$ 0.0070\\
    & & SGC& 0.8055 $\pm$ 0.0028& 0.5052 $\pm$ 0.0063\\
    & & GPRGNN &0.8685 $\pm$ 0.0023& 0.5009 $\pm$ 0.0035\\
    & \multirow{4}{*}{Chameleon} 
    & NLGCN& 0.6745 $\pm$ 0.0060& 0.4977 $\pm$ 0.0063\\
    & & NLGAT& 0.5965 $\pm$ 0.0069& 0.4849 $\pm$ 0.0066\\
    & & NLMLP& 0.4923 $\pm$ 0.0077& 0.3357 $\pm$ 0.0054\\
    & & GPRGNN &0.6541 $\pm$ 0.0064& 0.5145 $\pm$ 0.0053\\
    \bottomrule
    \end{tabular}}
\end{table*}

\begin{table*}[htbp]
  \label{tab:Ablation2}
  \resizebox{\textwidth}{!}{
    \begin{tabular}{lllll}
    \toprule
    Method & Dataset & GNN Models & Classify Acc &Attack AUROC \\
    \midrule
    \multirow{16}{*}{GTD (Two Stage \& Flattenning)} & \multirow{4}{*}{Cora} 
    & GCN& 0.8684 $\pm$ 0.0031& 0.4659 $\pm$ 0.0042\\
    & & GAT& 0.8657 $\pm$ 0.0018& 0.4522 $\pm$ 0.0056\\
    & & SGC& 0.8731 $\pm$ 0.0054& 0.4977 $\pm$ 0.0068\\
    & & GPRGNN &0.8806 $\pm$ 0.0030& 0.4894 $\pm$ 0.0058\\
    & \multirow{4}{*}{CiteSeer} 
    & GCN& 0.7233 $\pm$ 0.0035& 0.4035 $\pm$ 0.0061\\
    & & GAT& 0.7210 $\pm$ 0.0042& 0.3789 $\pm$ 0.0049\\
    & & SGC& 0.7362 $\pm$ 0.0065& 0.4499 $\pm$ 0.0069\\
    & & GPRGNN &0.7358 $\pm$ 0.0047& 0.4619 $\pm$ 0.0062\\
    & \multirow{4}{*}{PubMed} 
    & GCN& 0.8381 $\pm$ 0.0023& 0.4990 $\pm$ 0.0048\\
    & & GAT& 0.8400 $\pm$ 0.0028& 0.4911 $\pm$ 0.0061\\
    & & SGC& 0.8080 $\pm$ 0.0020& 0.5005 $\pm$ 0.0057\\
    & & GPRGNN &0.8553 $\pm$ 0.0014& 0.4967 $\pm$ 0.0034\\
    & \multirow{4}{*}{Chameleon} 
    & NLGCN& 0.6657 $\pm$ 0.0062& 0.4854 $\pm$ 0.0065\\
    & & NLGAT& 0.6585 $\pm$ 0.0070& 0.4602 $\pm$ 0.0063\\
    & & NLMLP& 0.4824 $\pm$ 0.0074& 0.3048 $\pm$ 0.0051\\
    & & GPRGNN &0.6550 $\pm$ 0.0058& 0.4936 $\pm$ 0.0049\\
    \bottomrule
    \end{tabular}}
\end{table*}

For each experiment, we repeated 5 times and presented the mean and standard deviation of the results. The complete results are showed in Table~\ref{tab:Ablation1}. Gradient Ascent, which is proposed by~\citet{chen2022relaxloss}, refers to periodically using gradient ascent to update the parameters of the target model when the average train loss falls below a predefined threshold. Our analysis about Table~\ref{tab:Ablation1} is as follow:

We first focus on the improvement of defense capability. It can be observed that two-stage (without flattening), flattening, and gradient ascent all enhance the defense capability of the target model compared to normal training. The effect of two-stage (without flattening) on reducing the AUROC of the attack model is the most pronounced, followed by flattening, while gradient ascent slightly reduces it. These results align with expectations because two-stage (without flattening) directly enables the model to learn the distribution of the testing data and flattening decrease the difference between the loss distributions of training and testing nodes. Surprisingly, gradient ascent hardly improves the model's defense capability, suggesting that our method's exclusion of gradient ascent is reasonable.

Then we focus on the decline of classification accuracy caused by these variants. From the results, it can be seen that two-stage (without flattening) and gradient ascent hardly lead to a decrease in classification accuracy, and flattening only results in a slight decline. These results are interpretable: two-stage (without flattening) only used testing data for an extra training; gradient ascent has a minimal impact on the model's defense capability, which also means that it hardly change the model; flattening slightly alters the model's mapping when using soft labels, so the classification accuracy decrease. However, the degree of decline caused by flattening is acceptable compared to its enhancement in defense capability. In summary, our GTD method (two stage with flattening) is the best.

\subsection{The Influences of Dataset Split Ratios}

Because changes in the ratio of the trainset can significantly impact the model's generalization ability, we investigated the influence of different split ratios on GTD's defense capability. We conducted experiments both on hard and weak setting and only used GCN model. Results of attack model's AUROC shows in Table~\ref{tab:Real}. For each experiment, we repeated 5 times and presented the mean and standard deviation of the results.

Table~\ref{tab:Real} lists five different trainset and testset split ratios. From the results, it can be observed that when the ratio is 9:1 and 3:1, the target models exhibit lower defense capability across all datasets. This is reasonable because with a larger split ratio of training data, the similarity between the shadow model and the target model is higher, making the target model more vulnerable to attacks. We can also see that the defense capability under the hard setting in the 9:1 split ratio is lower than that in the 3:1, which further corroborates the explanation. 

When the split ratio is 1:1, PubMed and Computers begin to exhibit a phenomenon where the weak setting is more susceptible to attacks. As the ratio of the trainset decreases further, Cora and Squirrel also show this trend. The reason for this phenomenon is that the reduction in the size of the training dataset leads to a deterioration in the imitation of the target model by the shadow model, thereby misleading the training of the attack model. However, under the weak setting, there is an intersection between the trainset of the shadow model and the testset of the target model. Therefore, when under attack, the shadow model has some knowledge about the testset of the target model, making it more vulnerable to attacks. 
\begin{table*}[htbp]
  \caption{Our defense method under different split ratios in Real Datasets}
  \label{tab:Real}
  \resizebox{\textwidth}{!}{
  \begin{tabular}{llllll}
    \toprule
    \multicolumn{2}{c}{\diagbox[width=8em]{Train:Test}{Dataset}} & Cora& CiteSeer & PubMed & Chameleon\\  
    \midrule
    \multirow{2}{*}{$9:1$} 
    & Hard & 0.505 $\pm$ 0.0028& 0.458 $\pm$ 0.0033& 0.505 $\pm$ 0.0031& 0.533 $\pm$ 0.0048\\
    & Weak & 0.490 $\pm$ 0.0045& 0.430 $\pm$ 0.0055& 0.490 $\pm$ 0.0055& 0.494 $\pm$ 0.0066\\
    \multirow{2}{*}{$3:1$} 
    & Hard & 0.483 $\pm$ 0.0035& 0.495 $\pm$ 0.0040& 0.498 $\pm$ 0.0039& 0.514 $\pm$ 0.0057\\
    & Weak & 0.476 $\pm$ 0.0056& 0.481 $\pm$ 0.0064& 0.500 $\pm$ 0.0063& 0.484 $\pm$ 0.0075\\
    \multirow{2}{*}{$1:1$} 
    & Hard & 0.466 $\pm$ 0.0042& 0.407 $\pm$ 0.0063& 0.496 $\pm$ 0.0051& 0.492 $\pm$ 0.0063\\
    & Weak & 0.421 $\pm$ 0.0068& 0.402 $\pm$ 0.0076& 0.502 $\pm$ 0.0072& 0.477 $\pm$ 0.0086\\
    \multirow{2}{*}{$1:3$} 
    & Hard & 0.507 $\pm$ 0.0059& 0.508 $\pm$ 0.0089& 0.500 $\pm$ 0.0081& 0.525 $\pm$ 0.0087\\
    & Weak & 0.501 $\pm$ 0.0076& 0.501 $\pm$ 0.0107& 0.507 $\pm$ 0.0102& 0.500 $\pm$ 0.0115\\
    \multirow{2}{*}{$1:9$} 
    & Hard & 0.495 $\pm$ 0.0074& 0.524 $\pm$ 0.0111& 0.496 $\pm$ 0.0102& 0.518 $\pm$ 0.0114\\
    & Weak & 0.503 $\pm$ 0.0101& 0.505 $\pm$ 0.0146& 0.509 $\pm$ 0.0144& 0.512 $\pm$ 0.0161\\
    \bottomrule
    \end{tabular}}
\end{table*}

\begin{table*}[htbp]
  \caption{Our defense method under different split ratios in Real Datasets}
  \label{tab:Real1}
  \resizebox{\textwidth}{!}{
  \begin{tabular}{lllllllllll}
    \toprule
    \multirow{2}{*}{\diagbox[width=8em]{Dataset}{Train:Test}}&\multicolumn{2}{c}{$9:1$} & \multicolumn{2}{c}{$3:1$} & \multicolumn{2}{c}{$1:1$} & \multicolumn{2}{c}{$1:3$} & \multicolumn{2}{c}{$1:9$}\\ \cline{2-11} 
    & Hard&Weak& Hard&Weak& Hard&Weak& Hard&Weak& Hard&Weak \\ 
    \midrule
    Cora& 0.505& 0.490& 0.483& 0.476& 0.466& 0.421& 0.507& 0.501& 0.495& 0.503 \\
    CiteSeer& 0.458& 0.430& 0.495& 0.481& 0.407& 0.402& 0.508& 0.501& 0.524& 0.505 \\
    PubMed& 0.505& 0.490& 0.498& 0.500& 0.496& 0.502& 0.500& 0.507& 0.496& 0.509 \\
    Computers& 0.505& 0.499& 0.503& 0.501& 0.496& 0.504& 0.500& 0.500& 0.495& 0.501 \\
    Photo& 0.496& 0.500& 0.502& 0.499& 0.508& 0.493& 0.507& 0.491& 0.509& 0.483 \\
    Chameleon& 0.533& 0.494& 0.514& 0.484& 0.492& 0.477& 0.525& 0.500& 0.518& 0.512 \\
    Squirrel& 0.508& 0.473& 0.495& 0.493& 0.480& 0.480& 0.501& 0.502& 0.519& 0.523 \\
    \bottomrule
    \end{tabular}}
\end{table*}

\begin{table*}[htbp]
  \caption{Hard/Weak Attack AUROC in Synthetic Dataset}
  \label{tab:Synthetic1}
  \resizebox{\textwidth}{!}{
  \begin{tabular}{lllllllllll}
    \toprule
    \multirow{2}{*}{\diagbox[width=7em]{$\phi$}{Train:Test}}&\multicolumn{2}{c}{$9:1$} & \multicolumn{2}{c}{$3:1$} & \multicolumn{2}{c}{$1:1$} & \multicolumn{2}{c}{$1:3$} & \multicolumn{2}{c}{$1:9$}\\ \cline{2-11} 
    & Hard&Weak& Hard&Weak& Hard&Weak& Hard&Weak& Hard&Weak \\ 
    \midrule
    -1& 0.550& 0.501& 0.513& 0.485& 0.509& 0.484& 0.511& 0.490& 0.519& 0.484\\
    -0.75& 0.541& 0.498& 0.510& 0.490& 0.504& 0.485& 0.516& 0.501& 0.522& 0.491\\
    -0.5& 0.531& 0.491& 0.510& 0.492& 0.510& 0.489& 0.515& 0.505& 0.527& 0.500\\
    -0.25& 0.518& 0.492& 0.513& 0.496& 0.515& 0.500& 0.514& 0.506& 0.490& 0.474 \\
    0& 0.533& 0.506& 0.513& 0.501& 0.501& 0.490& 0.500& 0.495& 0.492& 0.490\\
    0.25& 0.525& 0.503&0.508& 0.489& 0.505& 0.489&0.516& 0.500&0.490& 0.470 \\
    0.5 &0.531& 0.500& 0.503& 0.484& 0.503& 0.482& 0.514& 
    0.500& 0.530& 0.502\\
    0.75& 0.529& 0.494& 0.515& 0.490& 0.502& 0.481& 0.517& 0.502& 0.518& 0.493\\
    1& 0.538& 0.498& 0.527& 0.499& 0.517& 0.495& 0.520& 0.502& 0.510& 0.482\\
    \bottomrule
    \end{tabular}}
\end{table*}

\subsection{The Description of cSBM}
\label{cSBM}
In cSBMs, the node features are Gaussian random vectors, where the mean of the Gaussian depends on the community assignment. The difference of the means is controlled by a parameter $\mu$, while the difference of the edge densities in the communities and between the communities is controlled by a parameter $\lambda$. Hence $\mu$ and $\lambda$ capture the “relative informativeness” of node features and the graph topology, respectively. To fairly and continuously control the extent of information carried by the node features and graph topology, we introduce a parameter $\phi$ and use it to represent $\mu$ and $\lambda$: $\mu =\sqrt{\frac{n}{f} (1+\epsilon )} \times cos(\phi * \frac{\pi }{2})$, $\lambda=\sqrt{(1+\epsilon )} \times sin(\phi * \frac{\pi }{2})$, where $n$ denotes the number of nodes, $f$ denotes the dimension of the node feature vector, $\epsilon$ is a tolerance value. The setting $\phi = 0$ indicates that only node features are informative, while $|\phi| = 1$ indicates that only the graph topology is informative. Moreover, $\phi = 1$ corresponds to strongly homophilic graphs while $\phi = -1$ corresponds to strongly heterophilic graphs. In our experiments, we set $n=1000$, average degree per node is 20, $f=100$, $\phi = \{-1, -0.75, -0.5, -0.25, 0, 0.25, 0.5, 0.75, 1\}$, $\epsilon=15$, and use GCN model.


\section{Broader Impact}
\label{Broader Impact}
In addition to privacy violations, identifying whether certain high-profile influencers (nodes) are part of the training set can provide insights into the social dynamics and strategies of model providers and companies. For instance, an attacker could infer if the company's recommendations are biased towards or against certain influential users based on the attack results. Consequently, our proposed GTD can address this issue beyond privacy concerns, offering a broader application in safeguarding against such biases and maintaining the integrity of the recommendation systems. Furthermore, the defense of MIA can also inspire new designs for graph unlearning techniques~\citep{chien2022certified,chien2022efficient,pan2023unlearning}.

\end{document}